\documentclass[runningheads]{llncs}
\usepackage{graphicx}
\usepackage{comment}
\usepackage{amsmath,amssymb} 
\usepackage{color}

\usepackage{graphicx}
\usepackage{subcaption}
\usepackage{float}
\usepackage{caption}	
\usepackage{lscape}                                         

\usepackage[lined,ruled,linesnumbered]{algorithm2e}
\usepackage{xcolor}
\usepackage{tikz}
\usetikzlibrary{fit,calc}

\usepackage{booktabs}                   
\usepackage{multirow}
\usepackage{paralist}
\usepackage{enumitem}
\usepackage{colortbl}

\usepackage{bm}                          
\usepackage{epsfig}                      
\usepackage{graphicx}                  
\usepackage{mathtools}

\usepackage{color}

\usepackage{comment}

\usepackage{url}  
\usepackage[pagebackref=false,breaklinks=true,colorlinks=true,filecolor=blue,urlcolor=blue,linkcolor=red,bookmarks=false]{hyperref}
\usepackage[nocompress]{cite}

\usepackage{listings}

\usepackage{xspace}
\usepackage{setspace}

\usepackage{nicefrac}
\usepackage{microtype}
\usepackage[utf8]{inputenc} 
\usepackage[T1]{fontenc}    

\def\etal{et~al.}			  
\def\eg{e.g.,~}               

\DeclareMathAlphabet{\altmathcal}{OMS}{cmsy}{m}{n}

\def\p{\hspace{-0.2mm}(\hspace{-0.2mm}p\hspace{-0.2mm})}
\def\fn{\altmathcal{F}}
\def\ij{i \rightarrow j}
\def\ji{j \rightarrow i}
\def\Hji{\altmathcal{H}_{\!\ji}}
\def\Hij{\bm{H}_{\ij}}

\def\Ii{$\bm{I}_i$ }

\def\Mi{$\bm{M}_i$ }

\newlength\paramargin
\newlength\figmargin
\newlength\secmargin
\newlength\figcapmargin

\newcommand{\red}{\textcolor{red}}
\newcommand{\blue}{\textcolor{blue}}

\newcommand {\first}[1]{{\color{red}\textbf{#1}}}
\newcommand {\second}[1]{{\color{blue}\underline{#1}}}

\newcommand{\mpage}[2]
{
\begin{minipage}{#1\linewidth}\centering
#2
\end{minipage}
}

\newcommand{\heading}[1]
{
\vspace{1mm}\noindent\textbf{#1}
}

\newcommand{\secref}[1]{Section~\ref{sec:#1}}
\newcommand{\figref}[1]{Figure~\ref{fig:#1}} 
\newcommand{\tabref}[1]{Table~\ref{tab:#1}}

\long\def\ignorethis#1{}

\newcommand{\tb}[1]{\textbf{#1}}



\def\xi{\mathbf{x}_i}

\graphicspath{{figure}, {example}}

\usepackage[width=122mm,left=12mm,paperwidth=146mm,height=193mm,top=12mm,paperheight=217mm]{geometry}

\begin{document}
\pagestyle{headings}
\mainmatter
\def\ECCVSubNumber{1715}  

\title{Flow-edge Guided Video Completion}
\titlerunning{Flow-edge Guided Video Completion}

\author{
Chen Gao\inst{1} \and
Ayush Saraf\inst{2} \and
Jia-Bin Huang\inst{1} \and
Johannes Kopf\inst{2}
}
\authorrunning{C. Gao et al.}
%
\institute{${}^{1}$ Virginia Tech \quad ${}^{2}$ Facebook
}

\maketitle
\begin{center}
\centering
\newlength\fta
\setlength\fta{2.35cm}
\newlength\ftb
\setlength\ftb{0.8mm}
\newlength\ftc
\setlength\ftc{0.8mm}
\parbox[t]{\fta}{\centering%
  \includegraphics[width=\fta]{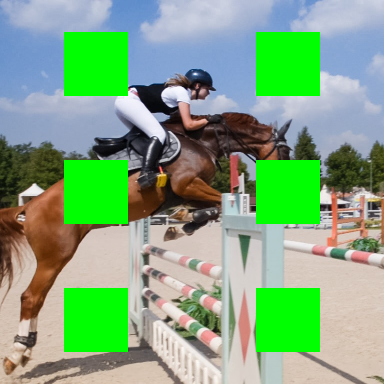}\\%
  \includegraphics[width=\fta]{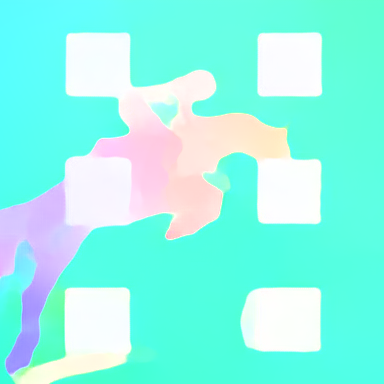}\\%
  \scriptsize (a) Input}%
\hfill%
\parbox[t]{\fta}{\centering%
  \includegraphics[width=\fta]{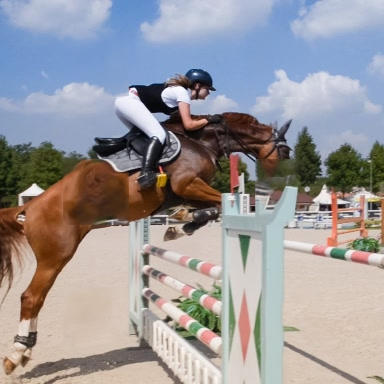}\\%
  \includegraphics[width=\fta]{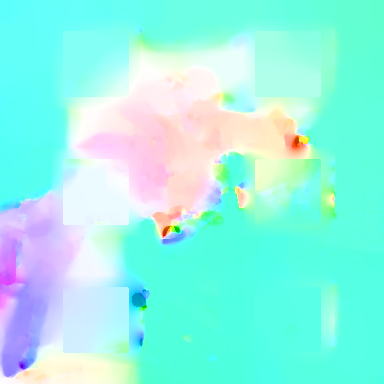}\\%
  \scriptsize (b) Huang et al.}%
%
\hfill%
\parbox[t]{\fta}{\centering%
  \includegraphics[width=\fta]{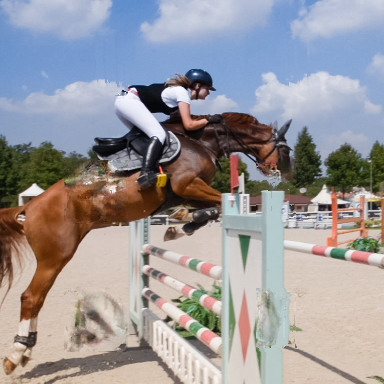}\\%
  \includegraphics[width=\fta]{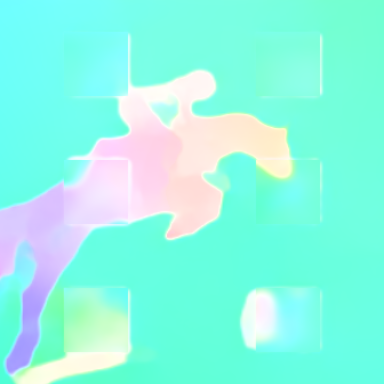}\\%
  \scriptsize (c) Xu et al.}%
%
\hfill%
\parbox[t]{\fta}{\centering%
  \includegraphics[width=\fta]{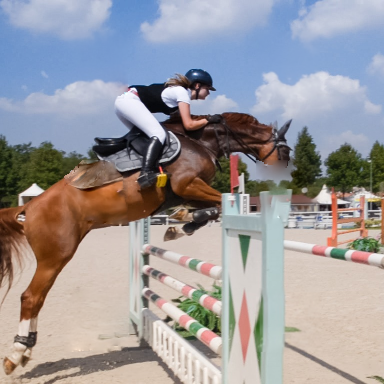}\\%
  \includegraphics[width=\fta]{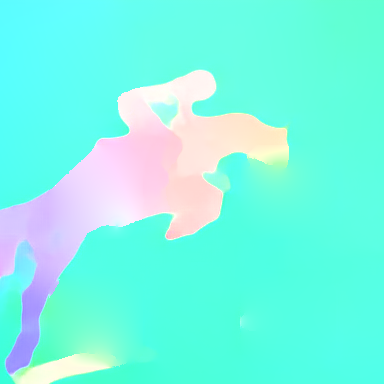}\\%
  \scriptsize (d) Our result}%
\hfill%
\parbox[t]{\fta}{\centering%
  \includegraphics[width=\fta]{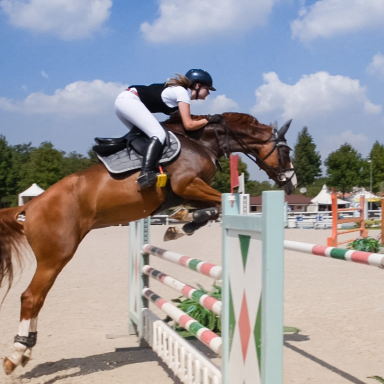}\\%
  \includegraphics[width=\fta]{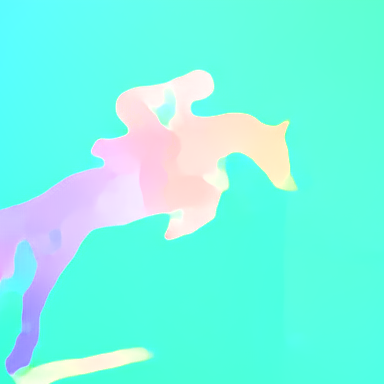}\\%
  \scriptsize (e) Ground truth}%
\captionof{figure}{\textbf{Flow-edge guided video completion.} 
Our new flow-based video completion method synthesizes sharper motion boundaries than previous methods and can propagate content across motion boundaries using non-local flow connections.
}
\label{fig:teaser}
\end{center}

\captionsetup{belowskip=-13pt}
\setlength{\abovedisplayskip}{5.5pt}
\setlength{\belowdisplayskip}{5.5pt}


\begin{abstract}
We present a new flow-based video completion algorithm.
Previous flow completion methods are often unable to retain the sharpness of motion boundaries. 
Our method first extracts and completes motion edges, and then uses them to guide piecewise-smooth flow completion with sharp edges.
Existing methods propagate colors among \emph{local} flow connections between adjacent frames.
However, not all missing regions in a video can be reached in this way because the motion boundaries form impenetrable barriers. 
Our method alleviates this problem by introducing \emph{non-local} flow connections to temporally distant frames, enabling propagating video content over motion boundaries.
We validate our approach on the DAVIS dataset. 
Both visual and quantitative results show that our method compares favorably against the state-of-the-art algorithms.
\end{abstract}

\section{Introduction}
\label{sec:intro}

\emph{Video completion} is the task of filling a given space-time region with newly synthesized content. 
It has many applications, including restoration (removing scratches), video editing and special effects workflows (removing unwanted objects), watermark and logo removal, and video stabilization (filling the exterior after shake removal instead of cropping).
The newly generated content should embed seamlessly in the video, and the alteration should be as imperceptible as possible. 
This is challenging because we need to ensure that the result is temporally coherent (does not flicker) and respects dynamic camera motion as well as complex object motion in the video.


Up until a few years ago, most methods used patch-based synthesis techniques \cite{Huang-SIGGRAPH-temporally,Newson-SIAM-Video,wexler2007space}. 
These methods are often slow and have limited ability to synthesize new content because they can only remix existing patches in the video.
Recent learning-based techniques achieve more plausible synthesis~\cite{wang2019video,Chang-ICCV-patchGAN}, but due to the high memory requirements of video, methods employing 3D spatial-temporal kernels suffer from resolution issues.
The most successful methods to date \cite{Huang-SIGGRAPH-temporally,Xu-CVPR-DFVI} are flow-based.
They synthesize color and flow jointly and propagate color along flow trajectories to improve temporal coherence, which alleviates memory problems and enables high-resolution output. 
Our method also follows this general approach.

The key to achieving good results with the flow-based approach is accurate flow completion, in particular, synthesizing sharp \emph{flow edges} along the object boundaries.
However, the aforementioned methods are not able to synthesize sharp flow edges and often produce over-smoothed results.
While this still works when removing \emph{entire} objects in front of \emph{flat} backgrounds, it breaks down in more complex situations.
For example, existing methods have difficulty in completing \emph{partially} seen \emph{dynamic} objects well (\figref{teaser}b--c).
Notably, this situation is ubiquitous when completing \emph{static screen-space masks}, such as logos or watermarks.
In this work, we improve the flow completion by explicitly completing flow edges. 
We then use the completed flow edges to guide the flow completion, resulting in \emph{piecewise-smooth} flow with sharp edges (\figref{teaser}d).

Another limitation of previous flow-based methods is that chained flow vectors between adjacent frames can only form \emph{continuous} temporal constraints.
This prevents constraining and propagating to many parts of a video.
For example, considering the situation of the periodic leg motion of a walking person:
here, the background is repeatedly visible between the legs, but the sweeping motion prevents forming continuous flow trajectories to reach (and fill) these areas.
We alleviate this problem by introducing additional flow constraints to a set of \emph{non-local} (i.e., temporally distant) frames.
This creates short-cuts across flow barriers and propagates color to more parts of the video.

Finally, previous flow-based methods propagate color values directly. 
However, the color often subtly changes over time in a video due to effects such as lighting changes, shadows, lens vignetting, auto exposure, and white balancing, which can lead to visible color seams when combining colors propagated from different frames.
Our method reduces this problem by operating in the gradient domain.

In summary, our method alleviates the limitations of existing flow-based video completion algorithms through the following key contributions:
\begin{enumerate}
\item \tb{Flow edges}: By explicitly completing flow edges, we obtain piecewise-smooth flow completion.
\item \tb{Non-local flow}: We handle regions that cannot be reached through transitive flow (e.g., periodic motion, such as walking) by leveraging non-local flow.
\item \tb{Seamless blending}: We avoid visible seams in our results through operating in the gradient domain.
\item \tb{Memory efficiency}: Our method handles videos with up to 4K resolution, while other methods fail due to excessive GPU memory requirements.
\end{enumerate}


We validate the contribution of individual components to our results and show clear improvement over the prior methods in both quantitative evaluation and the quality of visual results.

\begin{figure*}[t]
\includegraphics[width=\linewidth]{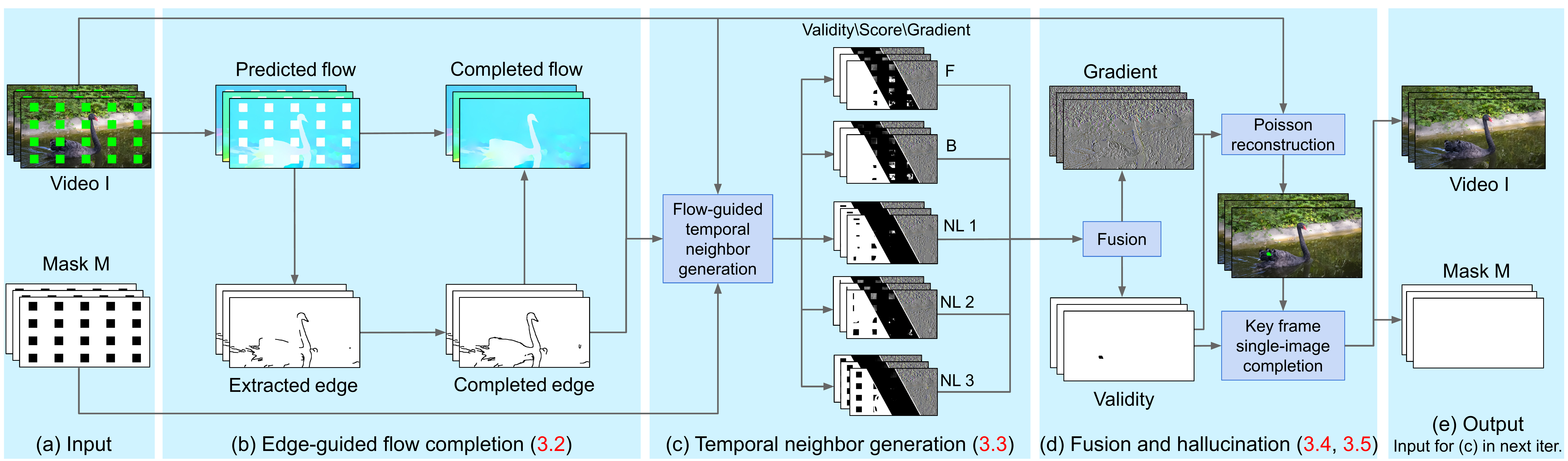}
\caption{
\textbf{Algorithm overview.} 
(a) The input to our video completion method is a color video and a binary mask video that indicates which parts need to be synthesized.
(b) We compute forward and backward flow between adjacent frames as well as a set of non-adjacent frames, extract and complete flow edges, and then use the completed edges to guide a piecewise-smooth flow completion (\secref{flow_completion}).
(c) We follow the flow trajectories to compute a set of candidate pixels for each missing pixel. For each candidate, we estimate a confidence score as well as a binary validity indicator (\secref{candidate_generation}).
(d) We fuse the candidates in the gradient domain for each missing pixel using a confidence-weighted average. We pick a frame with most missing pixels and fill it with image inpainting (\secref{candidate_fusion}). 
(e) The result will be passed into the next iteration until there is no missing pixel (\secref{Iterative}). 
}
\label{fig:overview}
\end{figure*}

\section{Related Work}
\label{sec:related}

\heading{Image completion} aims at filling missing regions in images with plausibly synthesized content.
\emph{Example-based} methods exploit the redundancy in natural images and transfer patches or segments from known regions to unknown (missing) regions \cite{criminisi2003object,drori2003fragment}.
These methods find correspondences for content transfer either via patch-based synthesis \cite{wexler2007space,barnes2009patchmatch} or by solving a labeling problem with graph cuts \cite{pritch2009shift,he2014image}.
In addition to using only verbatim copied patches, several methods improve the completion quality by augmenting patch search with geometric and photometric transformations\cite{mansfield2011transforming,darabi2012image,huang2013transformation,huang2014image}.
\emph{Learning-based} methods have shown promising results in image completion thanks to their ability to synthesize new content that may not exist in the original image \cite{pathak2016context,iizuka2017globally,Yan-ECCV-Shift,Yu2018-Generative}. 
Several improved architecture designs have been proposed to handle free-form holes \cite{Liu-ECCV-Partialconv,xie2019image,Yu-Gatedconv} and leverage predicted structures (e.g., edges) to guide the content \cite{Nazeri-ICCVW-Edgeconnect,xiong2019foreground,ren2019structureflow}.
Our work leverages a pre-trained image inpainting model \cite{Yu2018-Generative} to fill in pixels that are not filled through temporal propagation.



\heading{Video completion} inherits the challenges from the image completion problems and introduces new ones due to the additional time dimension.
Below, we only discuss the video completion methods that are most relevant to our work.
We refer the readers to a survey~\cite{ilan2015survey} for a complete map of the field.

Patch-based synthesis techniques have been applied to video completion by using 3D (spatio-temporal) patches as the synthesis unit~\cite{wexler2007space,Newson-SIAM-Video}.
It is, however, challenging to handle dynamic videos (e.g., captured with a hand-held camera) with 3D patches, because they cannot adapt to deformations induced by camera motion.
For this reason, several methods choose to fill the hole using 2D spatial patches and enforce temporal coherence with homography-based registration~\cite{granados2012background,gao2017augmented} or explicit flow constraints~\cite{strobel2014flow,roxas2014video,Huang-SIGGRAPH-temporally}.
In particular, Huang et al.~\cite{Huang-SIGGRAPH-temporally} propose an optimization formulation that alternates between optical flow estimation and flow-guided patch-based synthesis.
While the impressive results have been shown, the method is computationally expensive.
Recent work~\cite{bokov2018100+,okabe2019interactive} shows that the speed can be substantially improved by (1) decoupling the flow completion step from the color synthesis step and 
(2) removing patch-based synthesis (i.e., relying solely on flow-based color propagation).
These flow-based methods, however, are unable to infer sharp flow edges in the missing regions and thus have difficulties synthesizing dynamic object boundaries.
Our work focuses on overcoming the limitations of these flow-based methods. 
%
%
%
%
%

Driven by the success of learning-based methods for visual synthesis, recent efforts have focused on developing CNN-based approaches for video completion.
Several methods adopt 3D CNN architectures for extracting features and learning to reconstruct the missing content~\cite{wang2019video,Chang-ICCV-patchGAN}.
However, the use of 3D CNNs substantially limits the spatial (and temporal) resolution of the videos one can process due to the memory constraint.
To alleviate this issue, the methods in \cite{Kim-CVPR-VINet,Lee-ICCV-CPNet,Oh-ICCV-Onion} sample a small number of nearby frames as references. 
These methods, however, are unable to transfer temporally distant content due to the fixed temporal windows used by the method.
Inspired by flow-based methods~\cite{Huang-SIGGRAPH-temporally,bokov2018100+,okabe2019interactive}, Xu et al.~\cite{Xu-CVPR-DFVI} explicitly predict and complete dense flow field to facilitate propagating content from potentially distant frames to fill the missing regions.
Our method builds upon the flow-based video completion formulation and makes several technical contributions to substantially improve the visual quality of completion, including completing edge-preserving flow fields, leveraging non-local flow, and gradient-domain processing for seamless results.

\heading{Gradient-domain processing} techniques are indispensable tools for a wide variety of applications, including image editing~\cite{perez2003poisson,bhat2010gradientshop}, image-based rendering~\cite{kopf2013image}, blending stitched panorama~\cite{szeliski2011fast}, and seamlessly inserting moving objects in a video~\cite{chen2013motion}.
In the context of video completion, Poisson blending could be applied as a post-processing step to blend the synthesized content with the original video and hide the seams along the hole boundary.
However, such an approach would not be sufficient because the propagated content from multiple frames may introduce visible seams \emph{within} the hole that cannot be removed via Poisson blending.
Our method alleviates this issue by propagating gradients (instead of colors) in our flow-based propagation process.

\section{Method}
\label{sec:method}
%




\subsection{Overview}
\label{sec:overview}

The input to our video completion method is a color video and a binary mask video indicating which parts need to be synthesized (\figref{overview}a). 
We refer to the masked pixels as the \emph{missing} region and the others as the \emph{known} region.
Our method consists of the following three main steps.
\tb{(1) Flow completion}: 
We first compute forward and backward flow between adjacent frames as well as a set of non-adjacent (``non-local'') frames, and complete the missing region in these flow fields (\secref{flow_completion}).
Since edges are typically the most salient features in flow maps, we extract and complete them first.
We then use the completed edges to produce piecewise-smooth flow completion (\figref{overview}b).
\tb{(2) Temporal propagation}: 
Next, we follow the flow trajectories to propagate a set of candidate pixels for each missing pixel (\secref{candidate_generation}).
We obtain two candidates from chaining forward and backward flow vectors until a known pixel is reached.
We obtain three additional candidates by checking three temporally distant frames with the help of non-local flow vectors.
For each candidate, we estimate a confidence score as well as a binary validity indicator (\figref{overview}c).
\tb{(3) Fusion}:
We fuse the candidates for each missing pixel with at least one valid candidate using a confidence-weighted average (\secref{candidate_fusion}).
We perform the fusion in the gradient domain to avoid visible color seams (\figref{overview}d).

If there are still missing pixels after this procedure, it means that they could not be filled via temporal propagation (e.g., being occluded throughout the entire video).
To handle these pixels, we pick a single key frame (with most remaining missing pixels) and fill it completely using a single-image completion technique (\secref{Iterative}).
We use this result as input for another iteration of the same process described above.
The spatial completion step guarantees that we are making progress in each iteration, and its result will be propagated to the remainder of the video for enforcing temporal consistency in the next iteration.
In the following sections, we provide more details about each of these steps.

\subsection{Edge-guided Flow Completion}
\label{sec:flow_completion}

\begin{figure}[t]
\centering%
\newlength\ffca
\setlength\ffca{4cm} 
\newlength\ffcb
\setlength\ffcb{-0.0mm}
\parbox[t]{\ffca}{\centering%
  \fbox{\includegraphics[width=\ffca]{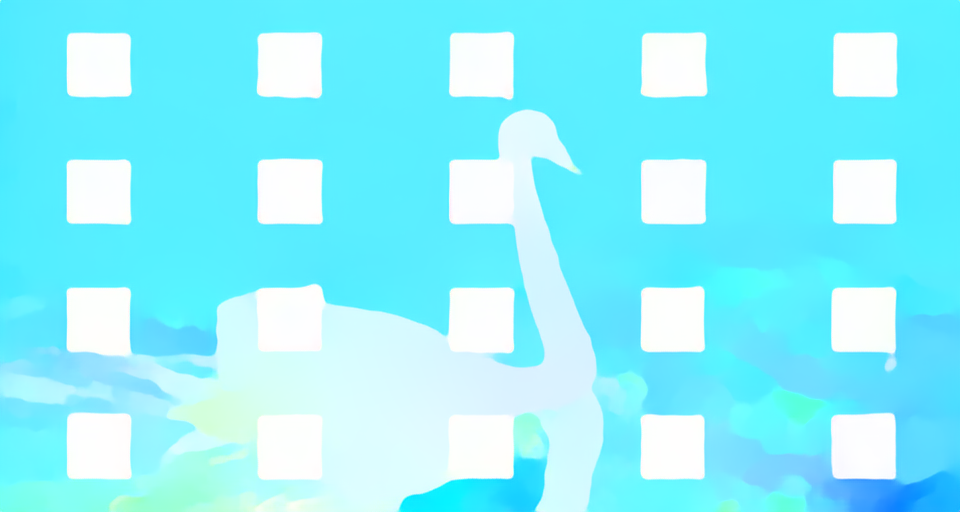}}\\%
	\scriptsize (a) Input flow}%
\hfill%
\parbox[t]{\ffca}{\centering%
  \fbox{\includegraphics[width=\ffca]{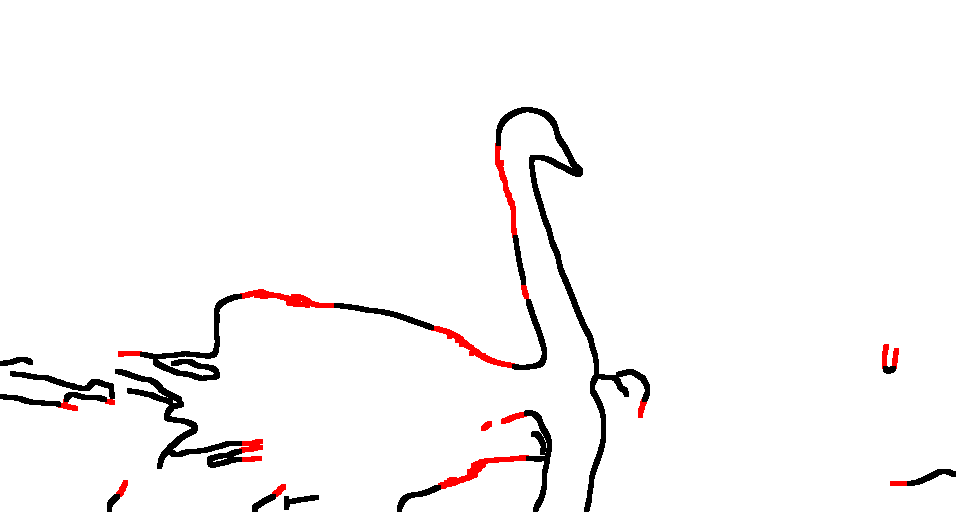}}\\%
	\scriptsize (b) Extracted/completed edges}%
\hfill%
\parbox[t]{\ffca}{\centering%
  \fbox{\includegraphics[width=\ffca]{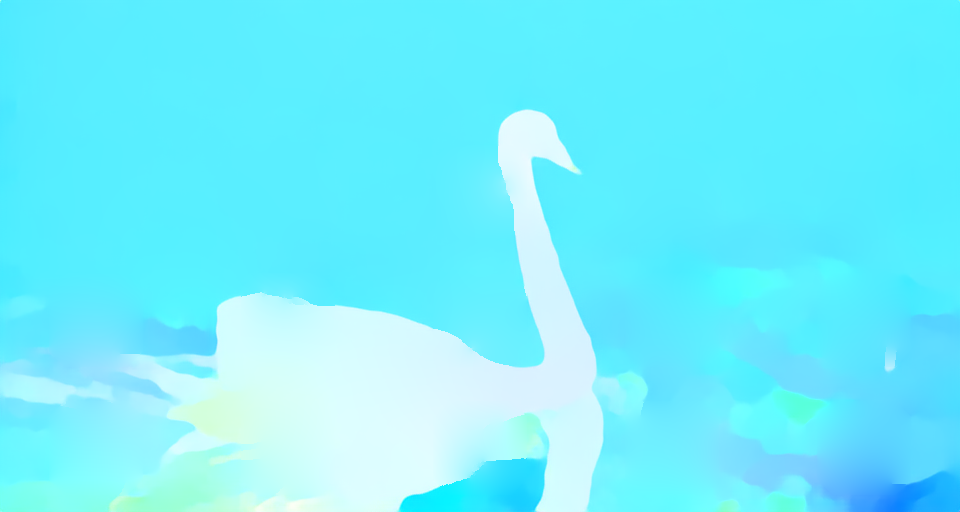}}\\%
	\scriptsize (c) Completed flow}%
	
\caption{
\textbf{Flow completion.}
(a) Optical flow estimated on the input video. Missing regions tend to have zero value (white).
(b) \textbf{Extracted} and \textbf{\red{completed}} flow edges.
(c) Piecewise-smooth completed flow, using the edges as guidance.}
\label{fig:flow_completion}
\end{figure}


The first step in our algorithm is to compute optical flow between adjacent frames as well as between several non-local frames (we explain how we choose the set of non-local connections in \secref{candidate_generation}) and to complete the missing regions in the flow fields in an edge-guided manner.

\heading{Flow computation.}
Let \Ii and \Mi be the color and mask of the $i$-th frame, respectively
(we drop the subscript $i$ if it is clear from the context),
with $\bm{M}\p = 1$ if pixel $p$ is missing, and $0$ otherwise.

We compute the flow between adjacent frames $i$ and $j$ using the pretrained FlowNet2 \cite{Ilg-CVPR-Flownet2} network~$\fn$:
\begin{equation}
\bm{F}_{i \rightarrow j} = 
\altmathcal{F} \big( \! \bm{I}_i,\,\bm{I}_j \! \big),
~~~
|i - j| = 1.
\label{eq:local_flow}
\end{equation}

Note that we set the missing pixels in the color video to black, but we do not treat them in any special way except during flow computation.
In these missing regions, the flow is typically estimated to be zero (white in visualizations, e.g., in \figref{flow_completion}a).

We notice that the flow estimation is \emph{substantially degraded} or even \emph{fails} in the presence of large motion, which frequently occurs in non-local frames.
To alleviate this problem, we use a homography warp $\Hji$ to compensate for the large motion between frame $i$ and frame $j$ (e.g., from camera rotation) before estimating the flow:
\begin{equation}
\bm{F}_{i \rightarrow j} = 
\altmathcal{F}
\big( \!
  \bm{I}_i,\,\Hji\!\left(\!\bm{I}_j\!\right)
\!\! \big)
+
\Hij,
~~~
|i - j| > 1.
\label{eq:homography_flow}
\end{equation}

Since we are not interested in the flow between the homography-aligned frames but between the original frames,
we add back the flow field $\Hij$ of the \emph{inverse} homography transformation, i.e., mapping each flow vector back to the original pixel location in the unaligned frame $j$.
We estimate the aligning homography using RANSAC on ORB feature matches \cite{Rublee-ICCV-ORB}. 
This operation takes about $3\%$ of the total computational time.

\heading{Flow edge completion.}
After estimating the flow fields, our next goal is to replace missing regions with plausible completions. 
We notice that the influence of missing regions extends slightly outside the masks (see the bulges in the white regions in \figref{flow_completion}a). 
Therefore, we dilate the masks by 15 pixels for flow completion.
As can be seen in numerous examples throughout this paper, flow fields are generally piecewise-smooth, i.e., their gradients are small except along distinct motion boundaries, which are the most salient features in these maps.
However, we observed that many prior flow-based video completion methods are unable to preserve sharp boundaries.
To improve this, we first extract and complete the flow edges, and then use them as guidance for a piecewise-smooth completion of the flow values.

We use the Canny edge detector \cite{Canny-1986:Edge} to extract a flow edge map $\bm{E}_{\ij}$ (\figref{flow_completion}b, black lines). Note that we remove the edges of missing regions using the masks.
We follow \emph{EdgeConnect} \cite{Nazeri-ICCVW-Edgeconnect} and train a flow edge completion network (See \secref{implementation} for details). 
At inference time, the network predicts a completed edge map $\tilde{\bm{E}}_{\ij}$ (\figref{flow_completion}b, red lines).

\heading{Flow completion.}
Now that we have hallucinated flow \emph{edges} in the missing region, we are ready to complete the actual flow \emph{values}.
Since we are interested in a smooth completion except at the edges, we solve for a solution that minimizes the gradients everywhere (except at the edges). We obtain the completed flow $\tilde{\bm{F}}$ by solving the following problem:
\begin{multline}
\underset{\tilde{\bm{F}}}{\operatorname{argmin}}
\sum_{p \mid \tilde{\bm{E}}(p)=1}
\big\| \Delta_x \tilde{\bm{F}}\p \big\|_2^2
\ +\ 
\big\| \Delta_y \tilde{\bm{F}}\p \big\|_2^2,
\\
\textrm{subject to}~~
\tilde{\bm{F}}\p = \bm{F}\p \mid \bm{M}\p = 0,
\label{eq:flow_completion}
\end{multline}
where $\Delta_x$ and $\Delta_y$ respectively denote the horizontal and vertical finite forward difference operator.
The summation is over all non-edge pixels, and the boundary condition ensures a smooth continuation of the flow outside the mask.
The solution to Equation~\ref{eq:flow_completion} is a set of sparse linear equations, which we solve using a standard linear least-squares solver.
\figref{flow_completion}c shows an example of flow completion.

\subsection{Local and Non-local Temporal Neighbors}
\label{sec:candidate_generation}
Now we can use the \emph{completed} flow fields to guide the completion of the color video. 
This proceeds in two steps: for each missing pixel, we
(1) find a set of known \emph{temporal neighbor} pixels (this section), and
(2) resolve a color by \emph{fusing} the candidates using weighted averaging (\secref{candidate_fusion}).


The flow fields establish a connection between related pixels across frames, which are leveraged to guide the completion by propagating colors from known pixels through the missing regions along flow trajectories. 
Instead of \emph{push}-propagating colors to the missing region (and suffering from repeated resampling), it is more desirable to transitively follow the forward and backward flow links for a given missing pixel, until known pixels are reached, and \emph{pull} their colors.

We check the validity of the flow by measuring the forward-backward cycle consistency error,
\begin{equation}
\tilde{\bm{D}}_{\ij}\p =
\Big\|
\bm{F}_{\ij}\p +
\bm{F}_{\ji}\hspace{-0.2mm}\big(\hspace{-0.2mm}p\!\,+\!\!\,\bm{F}_{\ij}\p\!\big)
\Big\|_2^2,
\label{eq:flow_error}
\end{equation}
and stop the tracing if we encounter an error of more than $\tau = 5$ pixels.
We call the known pixels that can be reached in this manner \emph{local} temporal neighbors because they are computed by chaining flow vector between adjacent frames.

\newlength\fttn
\setlength\fttn{2.15cm}
\newlength\fttm
\setlength\fttm{5.47cm}
\newlength\fttb
\setlength\fttb{0.0mm}

\begin{figure}[t]
\centering%
\parbox[t]{\fttn}{\centering%
  \includegraphics[width=\fttn]{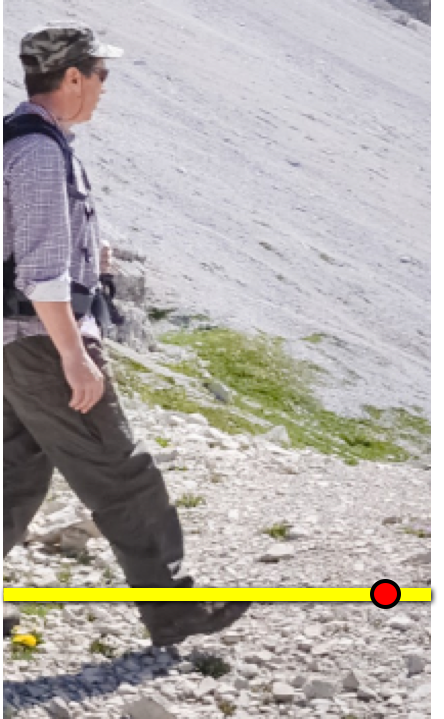}\\%
  \tiny Non-local frame 1}%
\hfill%
\parbox[t]{\fttn}{\centering%
  \includegraphics[width=\fttn]{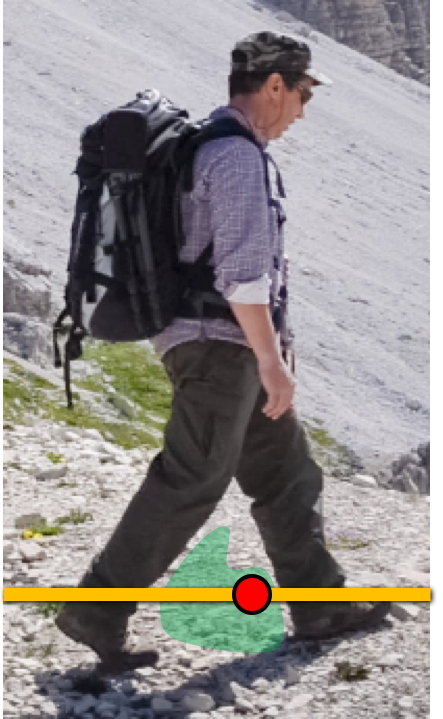}\\%
  \tiny Current frame}%
\hfill%
\parbox[t]{\fttn}{\centering%
  \includegraphics[width=\fttn]{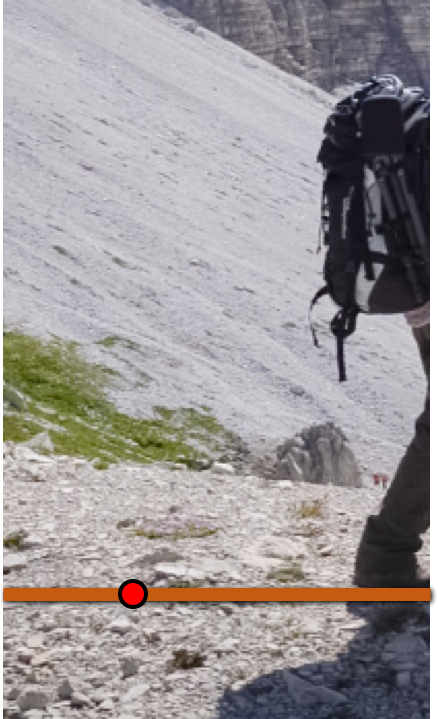}\\%
  \tiny Non-local frame 3}%
\hfill%
\parbox[t]{\fttm}{\centering%
  \includegraphics[width=\fttm,trim=0 90 0 90,clip]{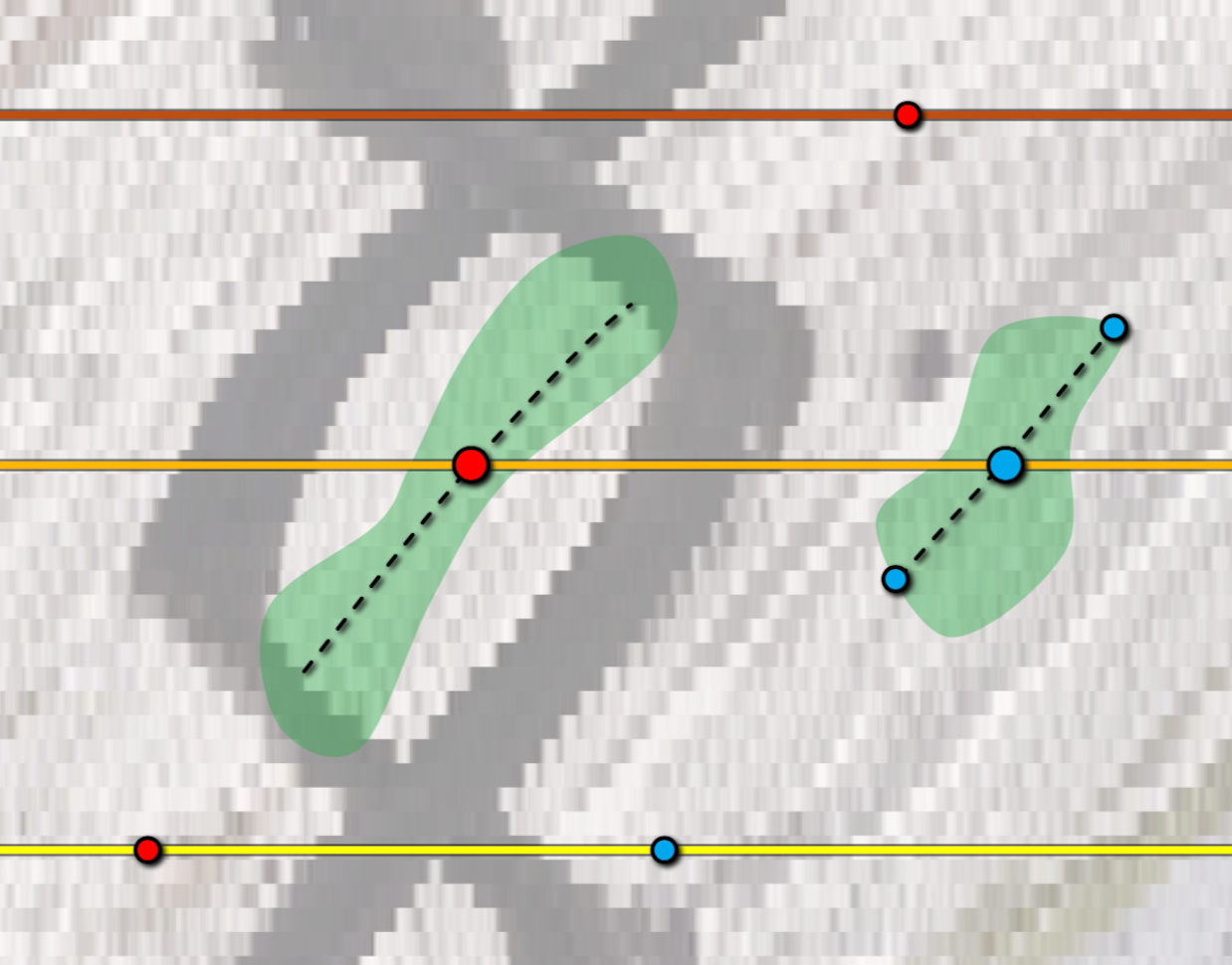}\\%
  \tiny space-time}%

%
%

\caption{
\textbf{Non-local completion candidates.}
The right figure shows a \emph{space-time} visualization for the highlighted scanlines in the left images.
Green regions are missing.
The yellow, orange and brown line in the right subfigure represents the scanline at the first non-local frame, the current frame and the third non-local frame, respectively. 
The figure illustrates the completion candidates for the \textbf{\red{red}} and \textbf{\blue{blue}} pixels (large discs on the orange line).
By following the flow trajectories (dashed black lines) until the edge of the missing region, we obtain \emph{local} candidates for the \textbf{\blue{blue}} pixel (small discs), but not for the \textbf{\red{red}} pixel, because the sweeping legs of the person form impassable flow barriers.
With the help of the non-local flow that connects to the temporally distant frames, we obtain extra \emph{non-local} neighbors for the \textbf{\red{red}} pixel (red discs on the yellow and brown line).
As a result, we can reveal the true background that is covered by the sweeping legs.
}
\label{fig:temporal_neighbor}
\end{figure}

Sometimes, we might not be able to reach a local known pixel, either because the missing region extends to the end of the video, because of invalid flow, or because we encounter a \emph{flow barrier}.
Flow barriers occur at every major motion boundary because the occlusion/dis-occlusion breaks the forward/backward cycle consistency there. 
A typical example is shown in \figref{temporal_neighbor}.
Barriers can lead to large regions of \emph{isolated} pixels without local temporal neighbors. Previous methods relied on hallucinations to generate content in these regions. 
However, hallucinations are more artifact-prone than propagation. 

In particular, even if the synthesized content is plausible, it will most likely be different from the actual content visible \emph{across} the barrier, which would lead to temporarily inconsistent results.

We alleviate this problem by introducing \emph{non-local} temporal neighbors, i.e., computing flow to a set of temporally distant frames that short-cut across flow barriers, which dramatically reduces the number of isolated pixels and the need for hallucination.
For every frame, we compute non-local flow to three additional frames using the homography-aligned method (Equation~\ref{eq:homography_flow}).
For simplicity, we always select the first, middle, and last frames of the video as non-local neighbors.
\figref{candidates} shows an example.

\heading{Discussion:}
We experimented with adaptive schemes for non-local neighbor selection, but found that the added complexity was hardly justified for the relatively short video sequences we worked with in this paper. When working with longer videos, it might be necessary to resort to more sophisticated schemes, such as constant frame offsets, and possibly adding additional non-local frames.

\subsection{Fusing Temporal Neighbors}
\label{sec:candidate_fusion}

\newlength\ftca
\setlength\ftca{2.4cm}
\newlength\ftcb
\setlength\ftcb{0.4mm}
\newlength\ftcc
\setlength\ftcc{-0.0mm}
\begin{figure}[t]
\centering%
\parbox[t]{\ftca}{\centering%
  \fbox{\includegraphics[width=\ftca]{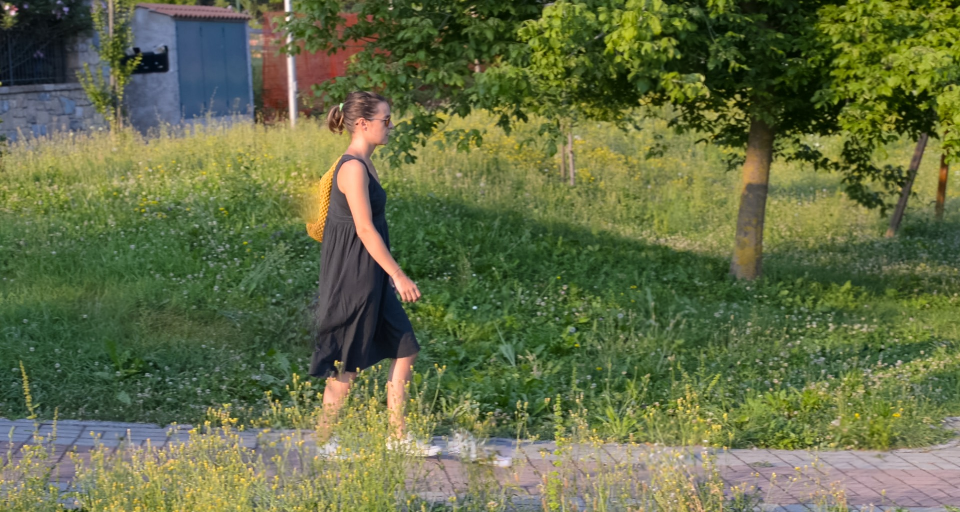}}\\%
  \fbox{\includegraphics[width=\ftca]{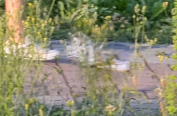}}\\%
	\tiny Forward neighbor}%
\hfill%
\parbox[t]{\ftca}{\centering%
  \fbox{\includegraphics[width=\ftca]{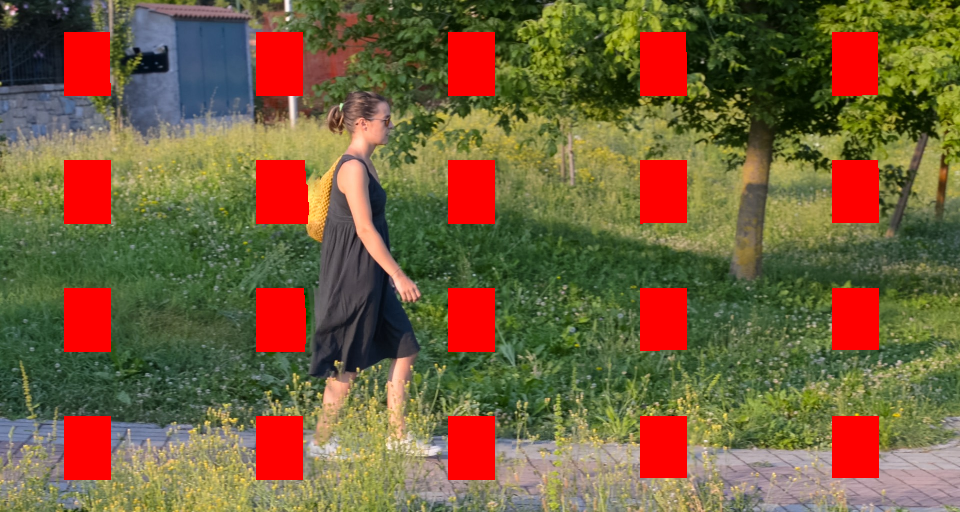}}\\%
  \fbox{\includegraphics[width=\ftca]{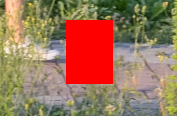}}\\%
	\tiny Backward neighbor}%
\hfill%
\parbox[t]{\ftca}{\centering%
  \fbox{\includegraphics[width=\ftca]{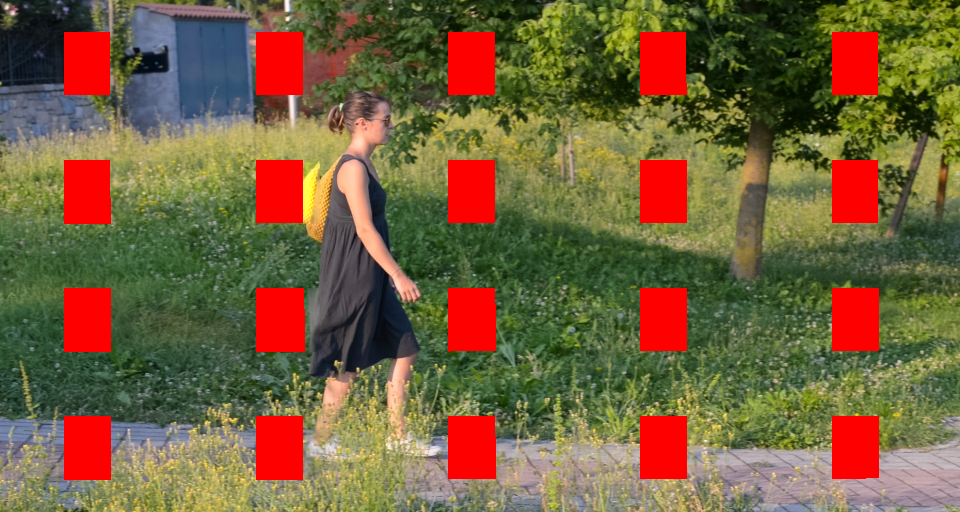}}\\%
  \fbox{\includegraphics[width=\ftca]{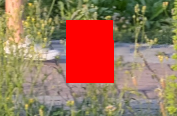}}\\%
	\tiny Non-local neighbor 1}%
\hfill%
\parbox[t]{\ftca}{\centering%
  \fbox{\includegraphics[width=\ftca]{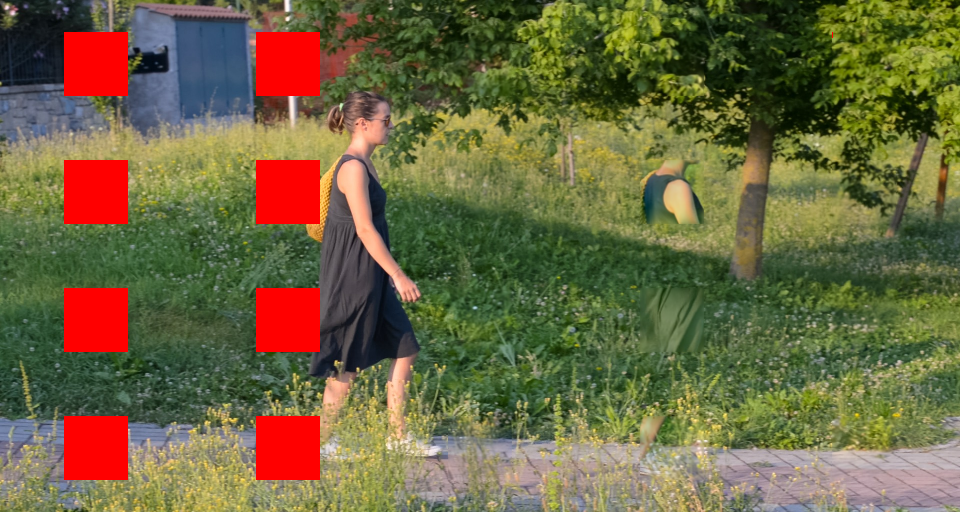}}\\%
  \fbox{\includegraphics[width=\ftca]{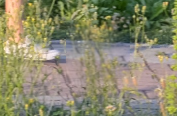}}\\%
	\tiny Non-local neighbor 2}%
\hfill%
\parbox[t]{\ftca}{\centering%
  \fbox{\includegraphics[width=\ftca]{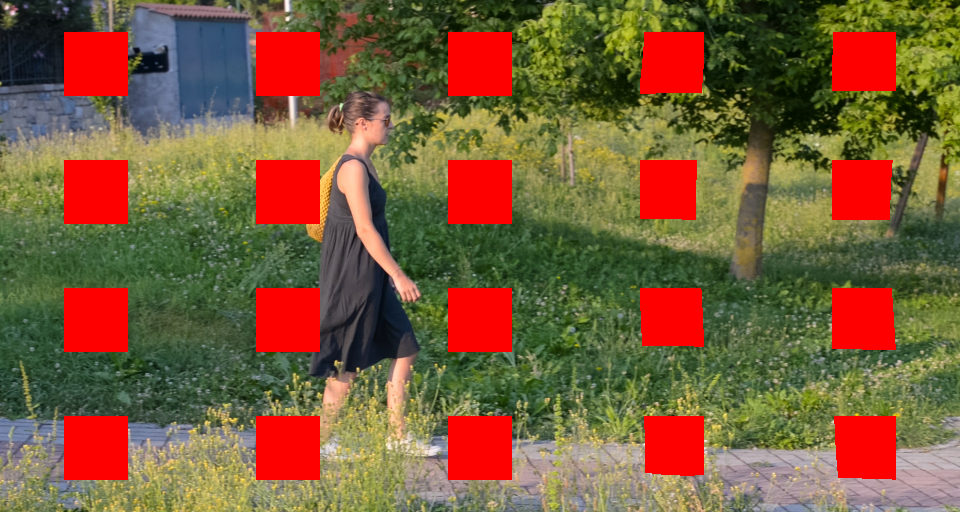}}\\%
  \fbox{\includegraphics[width=\ftca]{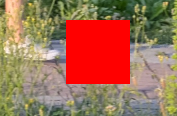}}\\%
	\tiny Non-local neighbor 3}%

\caption{\textbf{Temporal neighbors.}
Non-local temporal neighbors (the second non-local neighbor in this case) are useful when correct known pixels cannot be reached with local flow chains due to flow barriers 
(\textbf{\red{red}}: invalid neighbors.)
}
\label{fig:candidates}
\end{figure}

Now that we have computed temporal neighbors for the missing pixels, we are ready to fuse them to synthesize the completed color values. 
For a given missing pixel $p$, let $k \in N\p$ be the set of valid local and non-local temporal neighbors (we reject neighbors with flow error exceeding $\tau$, and will explain how to deal with pixels that have no neighbors in \secref{Iterative}).
We compute the completed color as a weighted average of the candidate colors $c_k$,
\begin{equation}
\tilde{\bm{I}}\p = \frac{\sum_k \! w_k c_k}{\sum_k \! w_k}.
\label{eq:fusion}
\end{equation}

The weights, $w_k$ are computed from the flow cycle consistency error:
\begin{equation}
w_k = \exp \! \left( -d_k / T \right),
\end{equation}
where $d_k$ is the consistency error $\tilde{\bm{D}}_{\ij}\p$ for \emph{non-local} neighbors, and the maximum of these errors along the chain of flow vectors for \emph{local} neighbors.
We set $T = 0.1$ to strongly down-weigh neighbors with large flow error.

\newlength\ftgd
\setlength\ftgd{4.0cm}
\newlength\ftge
\setlength\ftge{0.0mm}
\begin{figure}[t]

\centering%
\parbox[t]{\ftgd}{\centering%
\includegraphics[width=\ftgd]{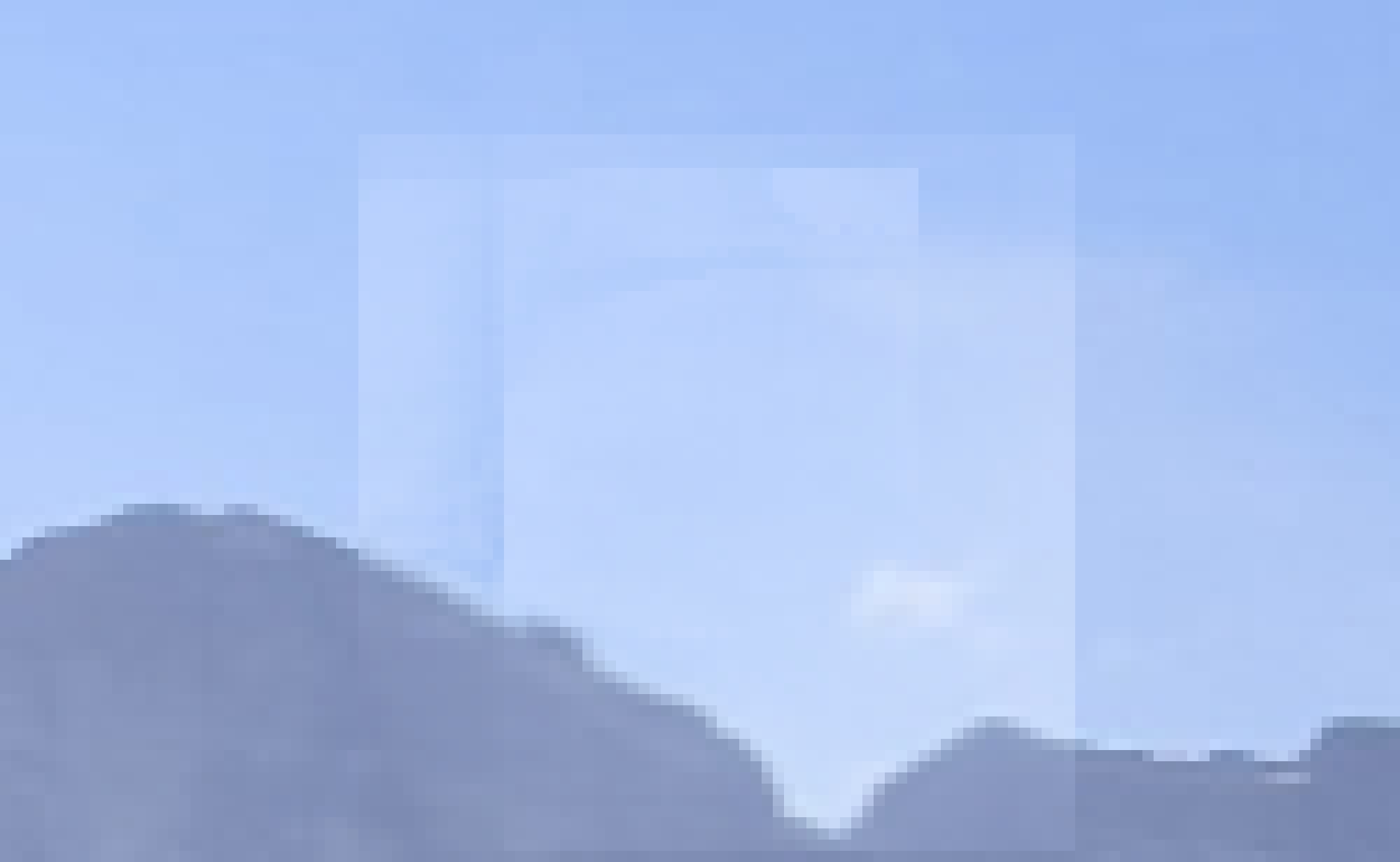}
\small (a) Color propagation}%
\hfill%
\parbox[t]{\ftgd}{\centering%
\includegraphics[width=\ftgd]{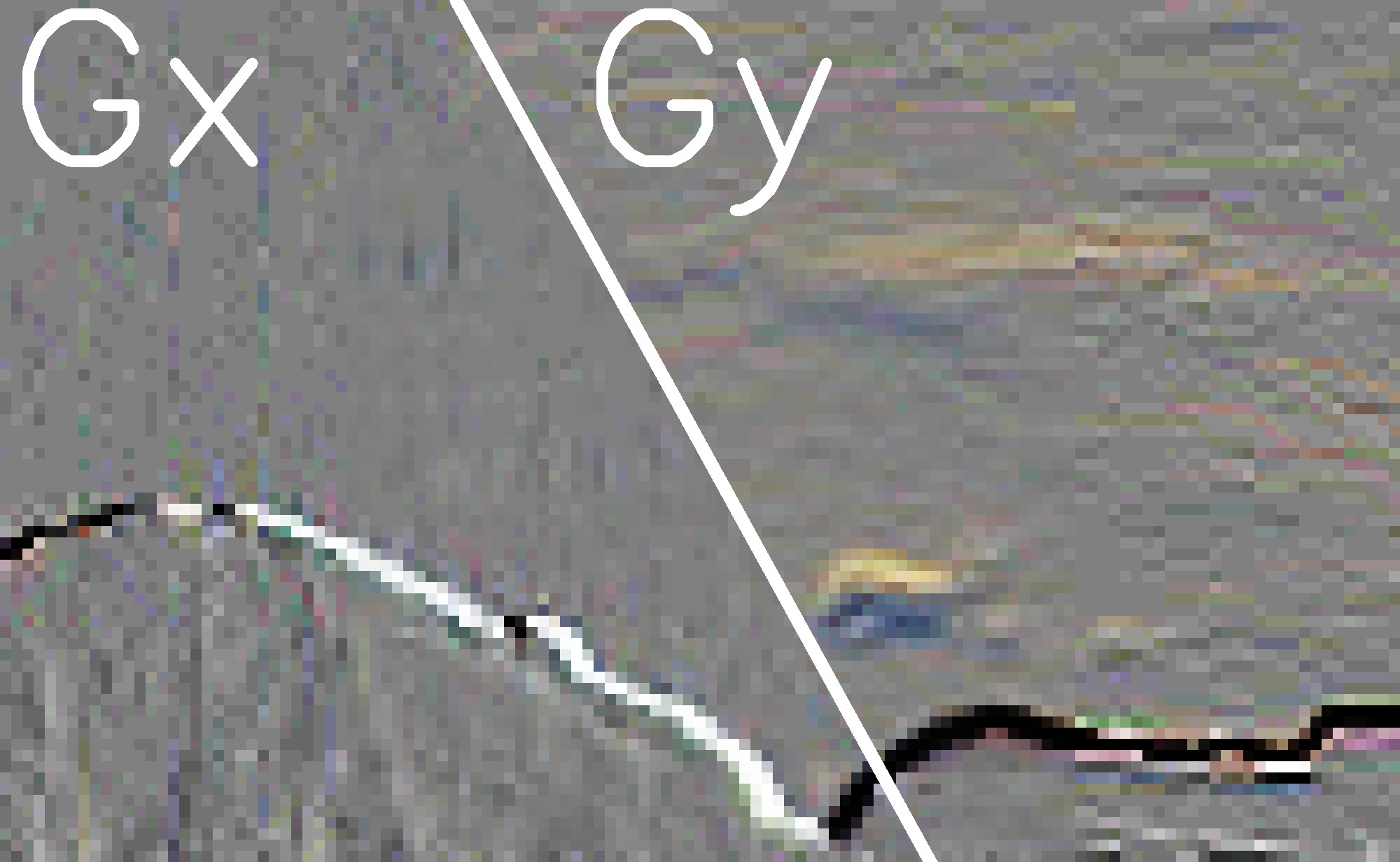}
\small (b) Propagated x/y gradient}%
\hfill%
\parbox[t]{\ftgd}{\centering%
\includegraphics[width=\ftgd]{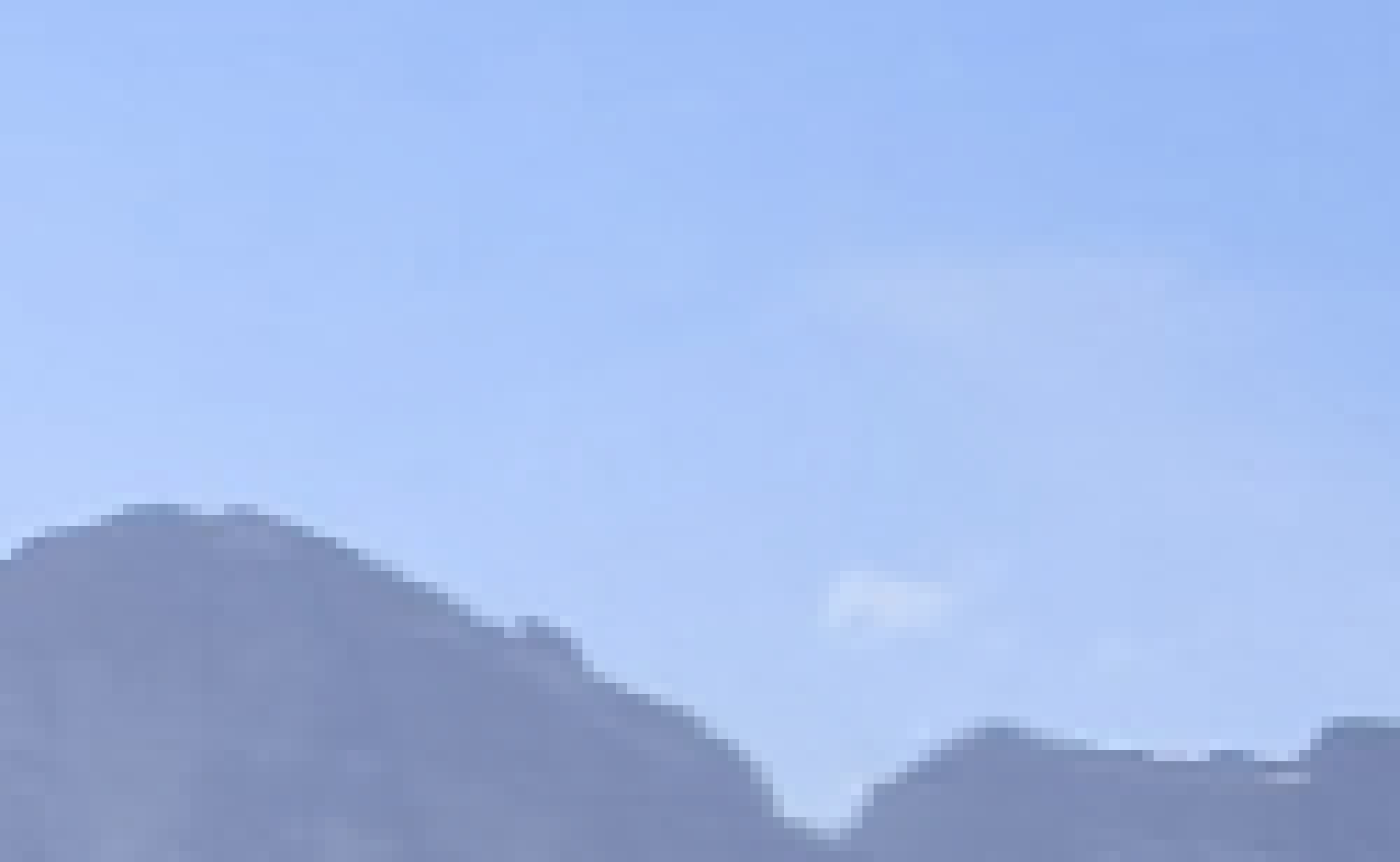}
\small (c) Reconstruction}%

\caption{\textbf{Gradient domain reconstruction.}
Previous methods operate directly in the color domain, which results in seams (a). We propagate in the gradient domain (b), and reconstruct the results by Poisson reconstruction (c).
}
\label{fig:gradient}
\end{figure}
\heading{Gradient-domain processing.}
We observed that directly propagating color values often yields visible seams, even with the correct flow. This is because of subtle color shifts in the input video (\figref{gradient}a). 
These frequently occur due to effects such as lighting changes, shadows, lens vignetting, auto exposure, and white balancing, etc.
We address this issue by changing Equation~\ref{eq:fusion} to compute a weighted average of color \emph{gradients}, rather than color values,
\begin{equation}
\tilde{\bm{G}_x}\p = \frac{\sum_k \! w_k \, \Delta_x c_k}{\sum_k \! w_k},~~~
\tilde{\bm{G}_y}\p = \frac{\sum_k \! w_k \, \Delta_y c_k}{\sum_k \! w_k},
\label{eq:fuse_gradient}
\end{equation}
and obtain the final image by solving a Poisson reconstruction problem,
\begin{multline}
\underset{\tilde{\bm{I}}}{\operatorname{argmin}}\,
\big\| \Delta_x \tilde{\bm{I}} - \tilde{\bm{G}_x} \big\|_2^2 + 
\big\| \Delta_y \tilde{\bm{I}} - \tilde{\bm{G}_y} \big\|_2^2,
\\
\textrm{subject to} \tilde{\bm{I}}\p = \bm{I}\p \mid M\p = 0,
\label{eq:poisson}
\end{multline}
which can be solved using a standard linear least-squares solver. 
By operating in the gradient domain (\figref{gradient}b), the color seams are suppressed (\figref{gradient}c).


\subsection{Iterative Completion}
\label{sec:Iterative}
In each iteration, we propagate color gradients and obtain (up to) five candidate gradients.
Then we fuse all candidate gradients and obtain missing pixel color values by solving a Poisson reconstruction problem (Equation~\ref{eq:poisson}). 
This will fill all the missing pixels that have valid temporal neighbors.
Some missing pixels might not have any valid temporal neighbors, even with the non-local flow, which, for example, happens when the pixel is occluded in all non-local frames, or when the flow is incorrectly estimated.
Similar to past work \cite{Huang-SIGGRAPH-temporally}, we formulate this problem as a single-image completion task, and solve it with Deepfill~\cite{Yu2018-Generative}.
However, if we would complete the remaining missing regions in \emph{all} frames with this single-image method, the result would not be temporally coherent.
Instead, we select only \emph{one} frame with the most remaining missing pixels and complete it with the single-image method.
Then, we feed the inpainting result as input to another iteration of our whole pipeline (with the notable exception of flow computation, which does not need to be recomputed).
In this subsequent iteration, the single-image completed frame is treated as a known region, and its color gradients are coherently propagated to the surrounding frames. 

The iterative completion process ends when there is no missing pixel.
In practice, our algorithm needs around 5 iterations to fill all missing pixels in the video sequences we have tried.
We have included the pseudo-code in the supplementary material, which summarizes the entire pipeline.



\section{Experimental Results}
\label{sec:results}

\subsection{Experimental setup}
\heading{Scenarios.}
We consider two application scenarios for video completion: (1) screen-space mask inpainting and (2) object removal.
For the inpainting setting, we generate a stationary mask with a uniform grid of $5 \times 4$ square blocks (see an example in \figref{qualitative_results}). 
This setting simulates the tasks of watermark or subtitle removal. 
Recovering content from such holes is particularly challenging because it often requires synthesizing \emph{partially} visible dynamic objects over their background.
For the object removal setting, we aim at recovering the missing content from a dynamically moving mask that covers the \emph{entire} foreground object. 
This task is relatively easier because, typically, the dominant dynamic object is removed entirely. 
Results in the object removal setting, however, are difficult to compare and evaluate due to the lack of ground truth content behind the masked object.
For this reason, we introduce a further \emph{synthetic} object mask inpainting task. 
Specifically, we take a collection of free-form object masks and randomly pair them with other videos, pretending there is an object occluding the scene.

\heading{Evaluation metrics.}
For tasks where the ground truth is available (stationary mask inpainting and object mask inpainting), we quantify the quality of the completed video using PSNR, SSIM, and LPIPS~\cite{zhang2018unreasonable}.
For LPIPS, we follow the default setting; we use Alexnet as the backbone, and we add a linear calibration on top of intermediate features.

\heading{Dataset.}
We evaluate our method on the DAVIS dataset \cite{Perazzi-CVPR-DAVIS}, which contains a total of 150 video sequences. 
Following the evaluation protocol in \cite{Xu-CVPR-DFVI}, we use the 60 sequences in \texttt{2017-test-dev} and \texttt{2017-test-challenge} for training our flow edge completion network. We use the 90 sequences in \texttt{2017-train} and \texttt{2017-val} for testing the stationary mask inpainting task. 
For the object removal task, we test on the 29 out of the 90 sequences for which refined masks provided by Huang \etal \cite{Huang-SIGGRAPH-temporally} are available (these masks include shadows cast by the foreground object). 
For the object mask inpainting task, we randomly pair these 29 video sequences with mask sequences from the same set that have the same or longer duration.
We resize the object masks by a uniform random factor in $\left[ 0.8,\ 1 \right]$, and trim them to match the number of frames. 
We resize all sequences to $960 \times 512$. 

\begin{table*}[t]
\caption{
\textbf{Video completion results with two types of synthetic masks.}
We report the average PSNR, SSIM and LPIPS results with comparisons to existing methods on DAVIS dataset. The best performance is in \first{bold} and the second best is \second{underscored}.
Missing entries indicate the method fails at the respective resolution.
}
\label{tab:results}
\centering
\resizebox{1\linewidth}{!} 
{
\begin{tabular}{l ccc c ccc c ccc c ccc}
\toprule
& \multicolumn{7}{c}{$720 \times 384$ resolution} && \multicolumn{7}{c}{$960 \times 512$ resolution}\\
\cline{2-8} \cline{10-16}
& 
\multicolumn{3}{c}{Stationary masks} && 
\multicolumn{3}{c}{Object masks} && 
\multicolumn{3}{c}{Stationary masks} && 
\multicolumn{3}{c}{Object masks} \\
\cline{2-4} \cline{6-8} \cline{10-12} \cline{14-16}
&PSNR $\uparrow$ & SSIM $\uparrow$ & LPIPS $\downarrow$ && PSNR $\uparrow$ & SSIM $\uparrow$ & LPIPS $\downarrow$ && PSNR $\uparrow$ & SSIM $\uparrow$ & LPIPS $\downarrow$ && PSNR $\uparrow$ & SSIM $\uparrow$ & LPIPS $\downarrow$ \\
	\midrule
Kim~\etal~\cite{Kim-CVPR-VINet}              & 25.19 & 0.8229 & 0.301 && 28.07 & 0.8673 & 0.283 && - & - & - && - & - & - \\
Newson~\etal~\cite{Newson-SIAM-Video}        & 27.50 & 0.9070 & \second{0.067} && 32.65 & 0.9648 & 0.023 && - & - & - && - & - & - \\
Xu~\etal~\cite{Xu-CVPR-DFVI}                 & 27.69 & 0.9264 & 0.077 && \second{39.67} & \second{0.9894} & \second{0.008} && 27.17 & \second{0.9216} & \second{0.085} && \second{38.88} & \second{0.9882} & \second{0.009} \\
Lee~\etal~\cite{Lee-ICCV-CPNet}              & 28.47 & 0.9170 & 0.111 && 35.76 & 0.9819 & 0.021 && \second{28.08} & 0.9141 & 0.117 && 35.34 & 0.9814 & 0.022 \\
Huang~\etal~\cite{Huang-SIGGRAPH-temporally} & 28.72 & 0.9256 & 0.070 && 34.64 & 0.9725 & 0.018 && - & - & - && - & - & - \\
Oh~\etal~\cite{Oh-ICCV-Onion}                & \second{30.28} & \second{0.9279} & 0.082 && 33.78 & 0.9630 & 0.058 && - & - & - && - & - & - \\
Ours                                         & \first{31.38} & \first{0.9592} & \first{0.042} && \first{42.72} & \first{0.9917} & \first{0.007} && \first{30.91} & \first{0.9564} & \first{0.048} && \first{41.89} & \first{0.9910} & \first{0.007} \\
\bottomrule
    \end{tabular}
    }
\end{table*}

\newlength\ftqa
\setlength\ftqa{2.0cm}
\newlength\ftqb
\setlength\ftqb{0.2mm}
\newlength\ftqc
\setlength\ftqc{0.8mm}

\begin{figure*}[t]

\includegraphics[width=\ftqa]{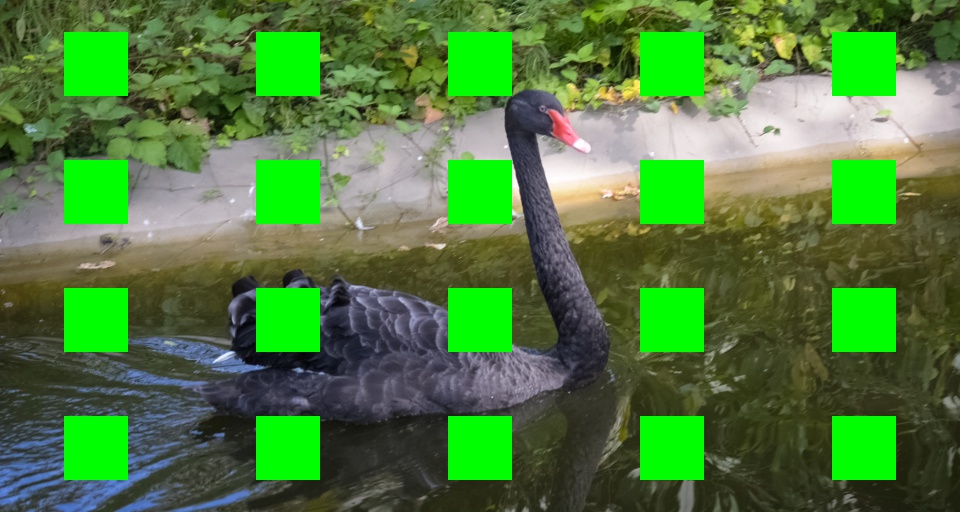}\hfill%
\includegraphics[width=\ftqa]{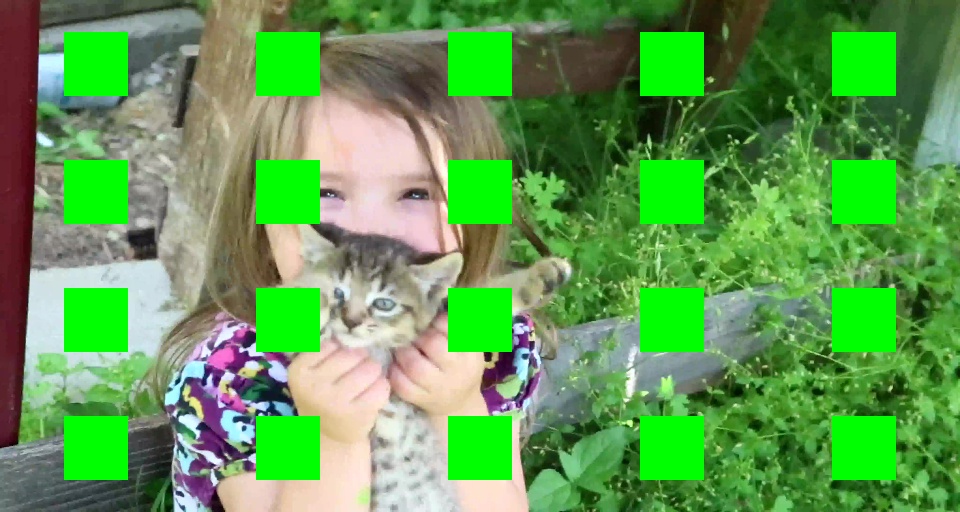}\hfill%
\includegraphics[width=\ftqa]{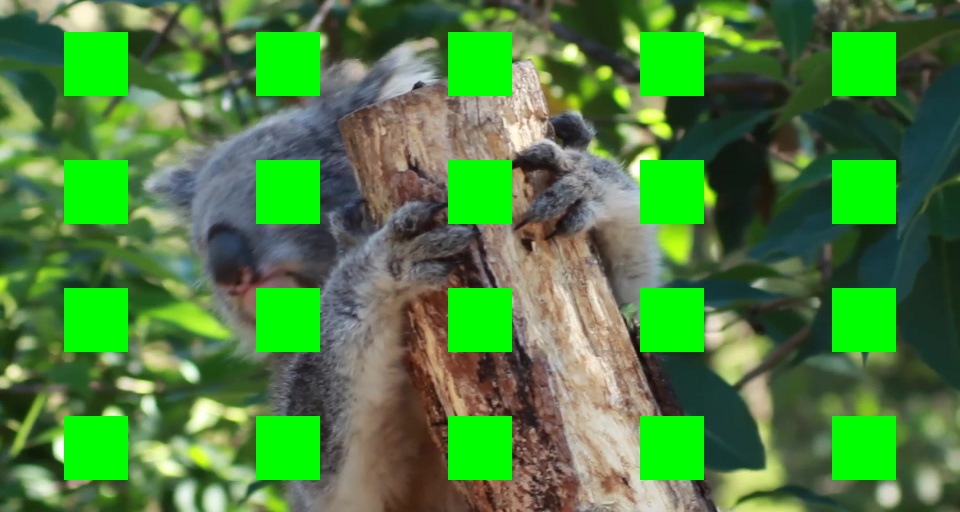}\hfill%
\includegraphics[width=\ftqa]{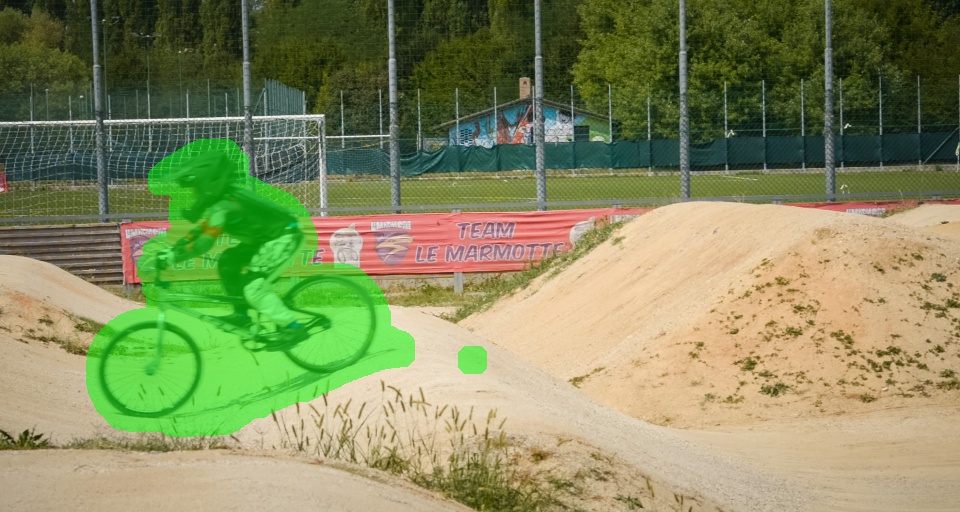}\hfill%
\includegraphics[width=\ftqa]{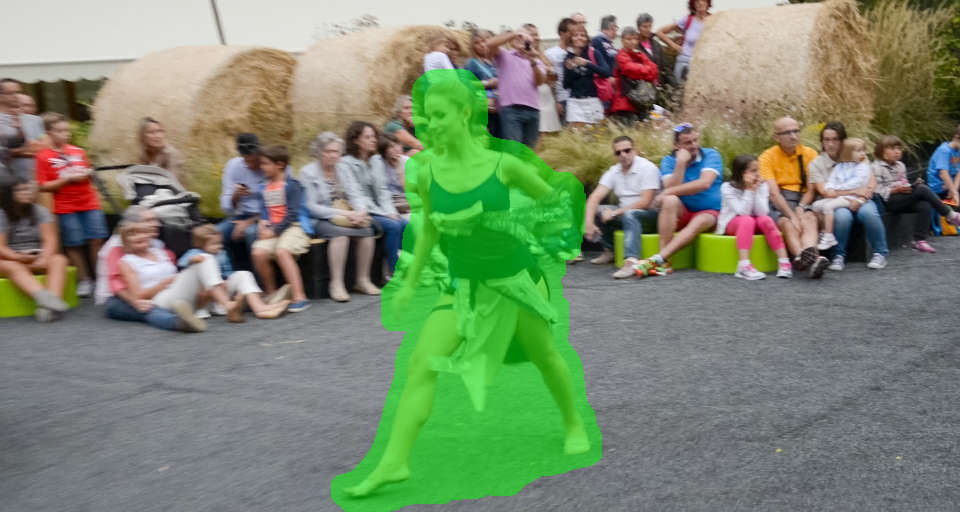}\hfill%
\includegraphics[width=\ftqa]{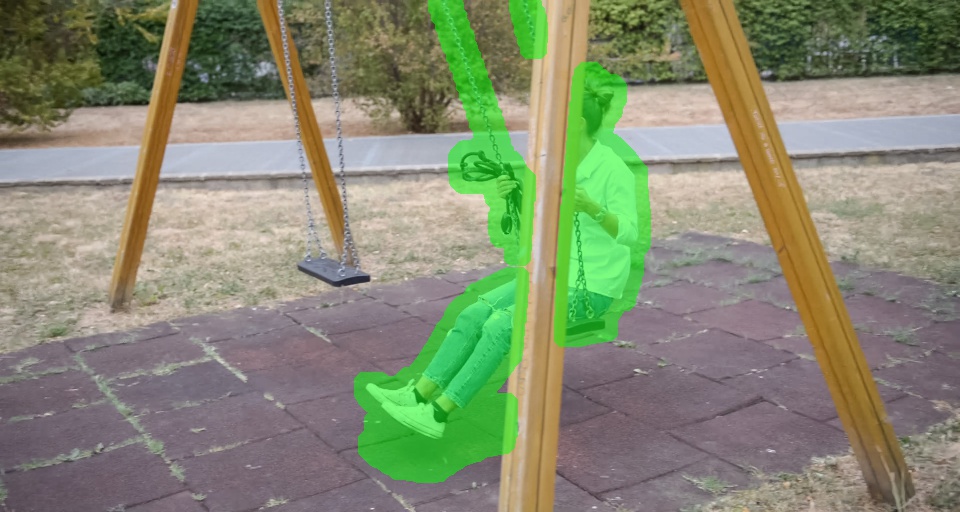}\\%
\includegraphics[width=\ftqa]{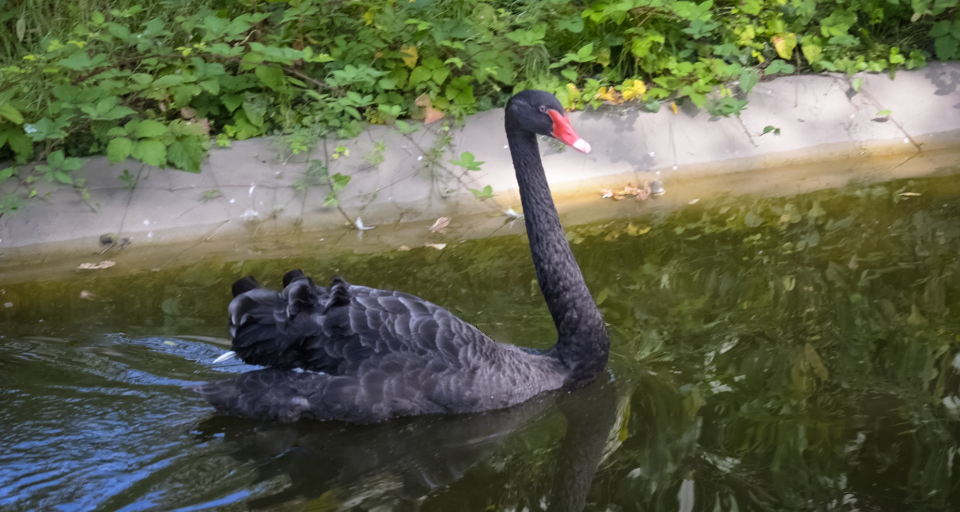}\hfill%
\includegraphics[width=\ftqa]{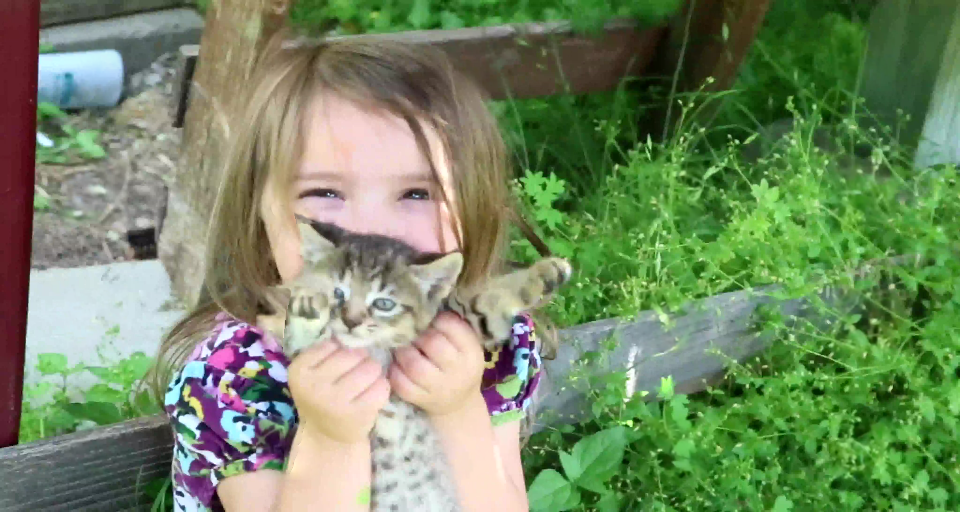}\hfill%
\includegraphics[width=\ftqa]{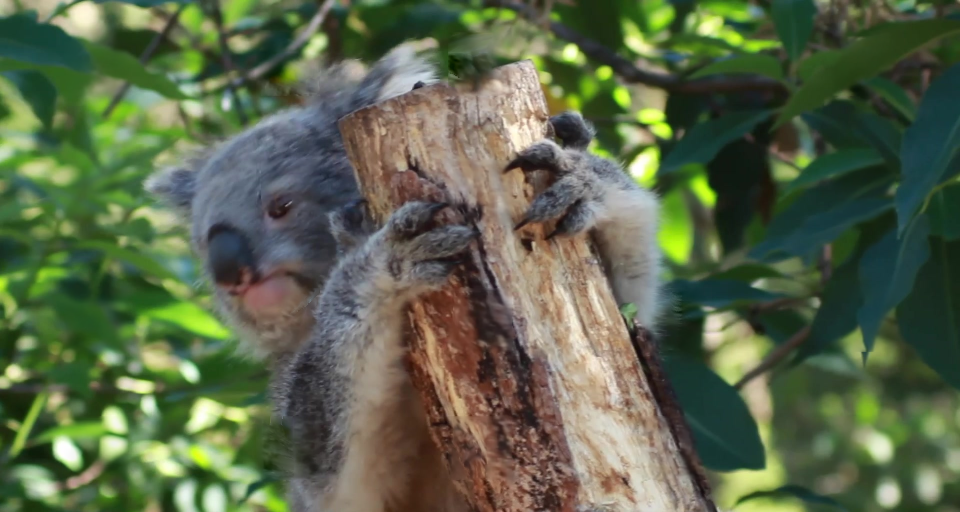}\hfill%
\includegraphics[width=\ftqa]{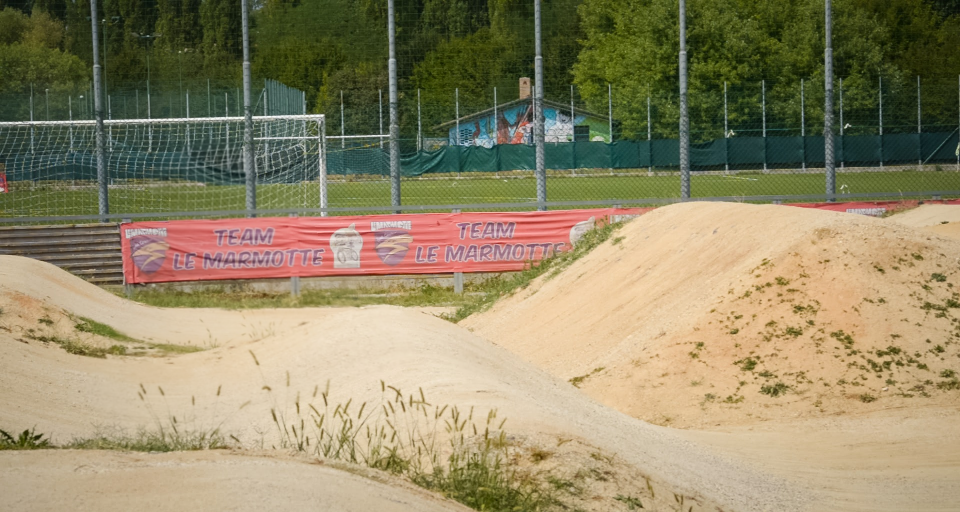}\hfill%
\includegraphics[width=\ftqa]{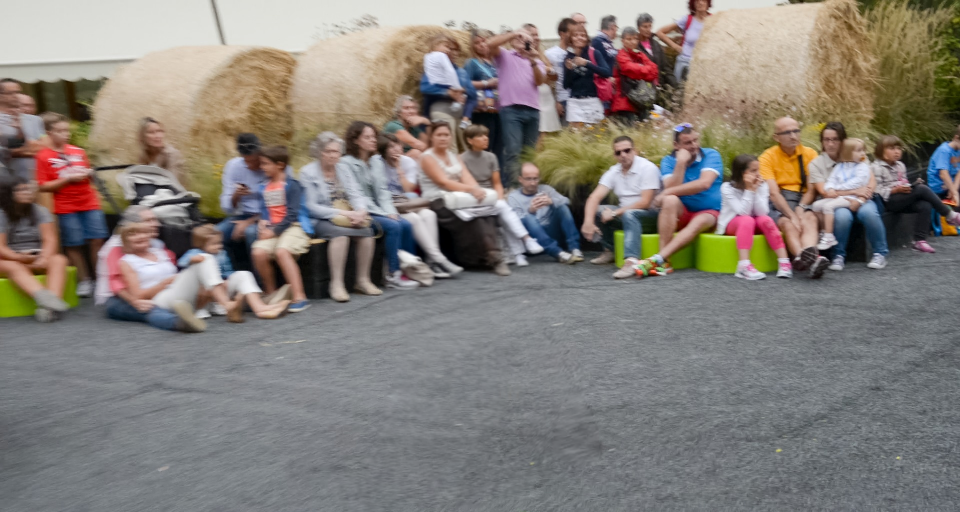}\hfill%
\includegraphics[width=\ftqa]{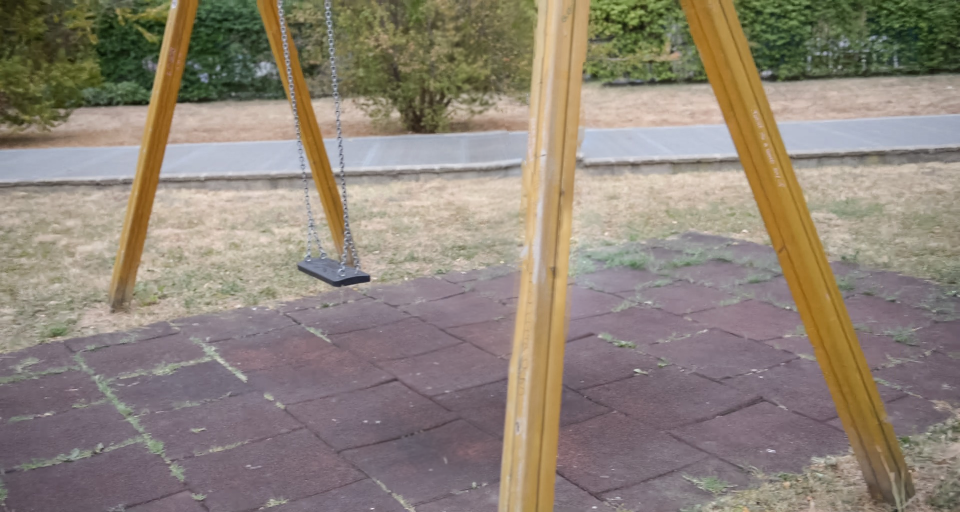}%
\caption{\textbf{Qualitative results.} We show the results of stationary screen-space inpainting task (first three columns) and object removal task (last three columns).
\label{fig:qualitative_results}
}
\end{figure*}

\heading{Implementation details.}
\label{sec:implementation}
We build our flow edge completion network upon the publicly available official implementation of EdgeConnect~\cite{Nazeri-ICCVW-Edgeconnect}\footnote{\url{https://github.com/knazeri/edge-connect}}.
We use the following parameters for the Canny edge detector~\cite{Canny-1986:Edge}: Gaussian $\sigma = 1$, low threshold $0.1$, high threshold $0.2$. We run the Canny edge detector on the flow magnitude image.
In addition to the mask and edge images, EdgeConnect takes a ``grayscale'' image as additional input; we substitute the flow magnitude image for it.
We load weights pretrained on the Places2 dataset \cite{zhou2017places}, and then finetune on 60 sequences in DAVIS \texttt{2017-test-dev} and \texttt{2017-test-challenge} for 3 epochs.
We adopt masks from NVIDIA Irregular Mask Dataset testing split \footnote{\url{https://www.dropbox.com/s/01dfayns9s0kevy/test_mask.zip?dl=0}}.
During training, we first crop the edge images and corresponding flow magnitude images to $256 \times 256$ patches. 
Then we corrupt them with a randomly chosen mask, which is resized to $256 \times 256$. 
We use the ADAM optimizer with a learning rate of $0.001$. 
Training our network takes 12 hours on a single NVIDIA P100 GPU.

\subsection{Quantitative evaluation}
\label{sec:quantitative}

We report quantitative results under the stationary mask inpainting and object mask inpainting setting in~\tabref{results}. 
Because not all methods were able to handle the full $960 \times 512$ resolution due to memory constraint, we downscaled all scenes to $720 \times 384$ and reported numbers for both resolutions. 
Our method substantially improves the performance over state-of-the-art algorithms~\cite{Oh-ICCV-Onion,Huang-SIGGRAPH-temporally,Lee-ICCV-CPNet,Xu-CVPR-DFVI,Newson-SIAM-Video,Kim-CVPR-VINet} on the three metrics.
Following~\cite{Huang-SIGGRAPH-temporally}, we also show the detailed running time analysis of our method in the supplementary material.
We report the time for each component of our method on the ``CAMEL'' video sequence under the object removal setting. 
Our method runs at $7.2$ frames per minute.

\subsection{Qualitative evaluation}
\figref{qualitative_results} shows sample completion results for a diverse set of sequences.
In all these cases, our method produces temporally coherent and visually plausible content.
Please refer to the supplementary video results for extensive qualitative comparison to the methods listed in \tabref{results}.

\subsection{Ablation study}
\label{sec:ablation}

\begin{figure}[t]
\centering%
\newlength\ftfc
\setlength\ftfc{3.0cm}
\newlength\ftfd
\setlength\ftfd{0.8mm}
\parbox[t]{\ftfc}{
\centering%
 \fbox{\includegraphics[width=\ftfc]{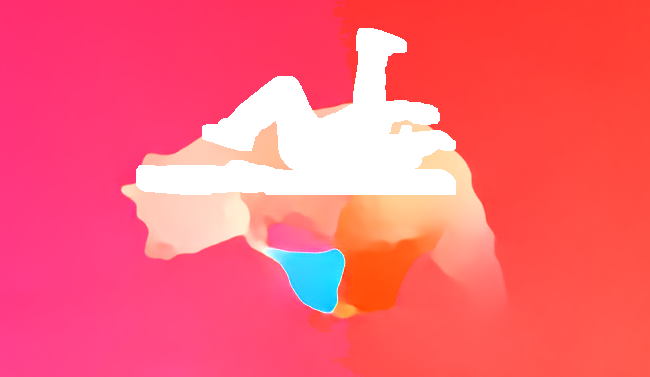}}\\%
	\footnotesize Input}%
\hfill%
\parbox[t]{\ftfc}{
\centering%
 \fbox{\includegraphics[width=\ftfc]{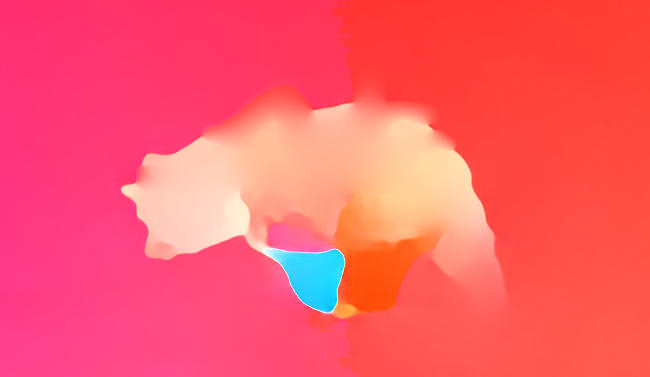}}\\%
	\footnotesize Diffusion}%
\hfill%
\parbox[t]{\ftfc}{
\centering%
 \fbox{\includegraphics[width=\ftfc]{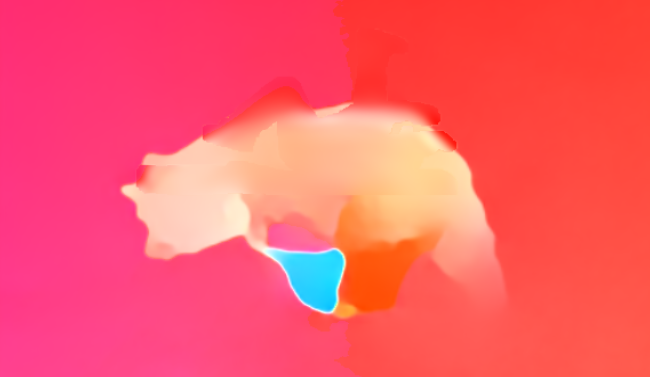}}\\%
	\footnotesize Xu~\etal~\cite{Xu-CVPR-DFVI}}%
\hfill%
\parbox[t]{\ftfc}{
\centering%
 \fbox{\includegraphics[width=\ftfc]{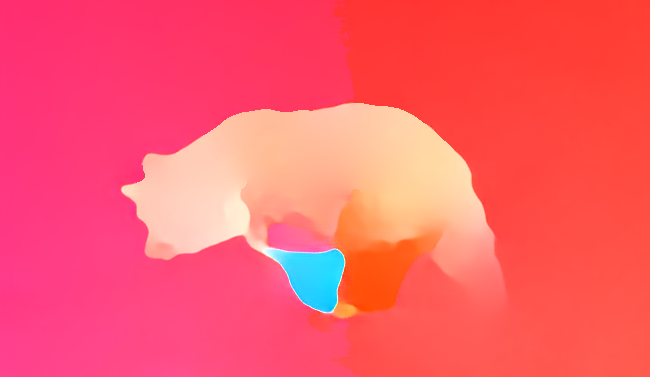}}\\%
	\footnotesize Ours}%
\caption{\textbf{Flow completion.} Comparing different methods for flow completion. Our method has better ability to retain the piecewise-smooth nature of flow fields (sharp motion boundaries, smooth everywhere else) than the other two methods.
}
\label{fig:flow}
\end{figure}
\newlength\ftgf
\setlength\ftgf{6.05cm}
\newlength\ftgg
\setlength\ftgg{0.0mm}
\begin{figure}[t]
\centering%
\parbox[t]{\ftgf}{\centering%
\includegraphics[width=\ftgf]{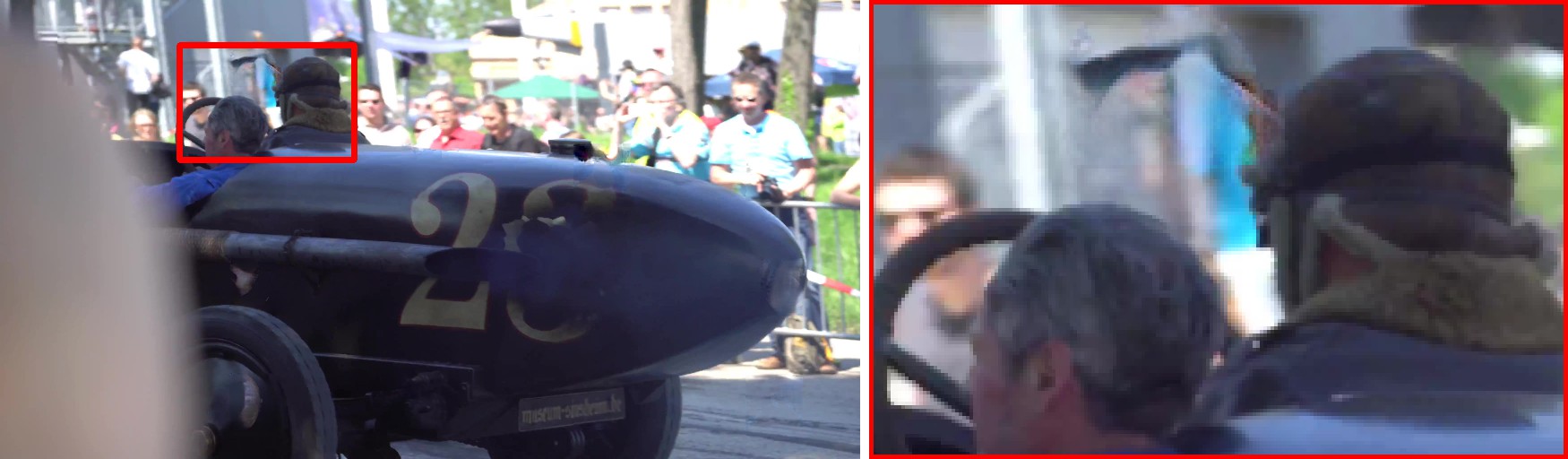}\\%
Without non-local neighbors}%
\hfill%
\parbox[t]{\ftgf}{\centering%
\includegraphics[width=\ftgf]{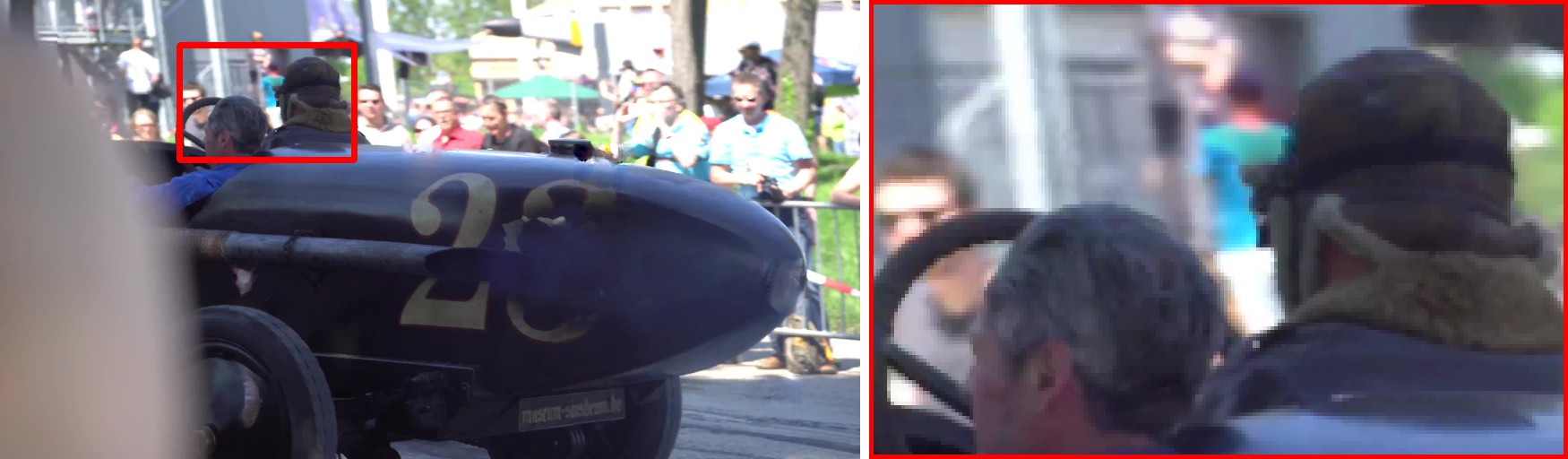}\\%
With non-local neighbors}%
\caption{\textbf{Non-local temporal neighbor ablation.}
Video completion results \emph{with} and \emph{without} non-local temporal neighbors.
The result without non-local neighbors (left) does not recover well from the lack of well-propagated content.
}
\label{fig:nonlocal}
\end{figure}

In this section, we validate the effectiveness of our design choices.

\heading{Gradient domain processing.} 
\begin{table}[t]
    \caption{
    \textbf{Ablation study.}
    We report the average scores on DAVIS.
    }
    \label{tab:ablation}
    \centering
    
    \begin{minipage}{0.45\linewidth}
    \centering
    {(a) \textbf{Domain and non-local}
    } 
    \end{minipage}
    \hfill
    \begin{minipage}{0.45\linewidth}
    {(b) \textbf{Flow completion methods} 
    }
    \end{minipage}
    \hfill
    \mpage{0.43}{
    \resizebox{1\linewidth}{!} 
    {
    \begin{tabular}{ccc | ccc c ccc}
    \toprule
    \multirow{2}{*}{Gradient} && \multirow{2}{*}{Non-local}  & \multicolumn{3}{c}{Stationary masks} && \multicolumn{3}{c}{Object masks } \\
    \cline{4-6} \cline{8-10}
    & & & PSNR $\uparrow$ & SSIM $\uparrow$ & LPIPS $\downarrow$ && PSNR $\uparrow$ & SSIM $\uparrow$ & LPIPS $\downarrow$ \\
    \midrule
         -     &&      -     & 28.28 & 0.9451 & 0.067 && 39.29 & 0.9893	& 0.009 \\
         -     && \checkmark & 28.47 & 0.9469 & 0.069 && 39.67 & 0.9897 & 0.009 \\
    \checkmark &&      -     & 30.78 & 0.9552 & 0.049 && 41.55 & 0.9907 & 0.007 \\
    \checkmark && \checkmark & 30.91 & 0.9564 & 0.048 && 41.89 & 0.9910 & 0.007 \\
    \bottomrule
    \end{tabular}
    }
}
\hfill
\mpage{0.53}{
\centering
    \resizebox{1\linewidth}{!} 
    {
    \begin{tabular}{l cccc c cccc}
    \toprule
    & \multicolumn{4}{c}{Stationary masks} && \multicolumn{4}{c}{Object masks} \\
    \cline{2-5} \cline{7-10}
    & Flow EPE $\downarrow$ & PSNR $\uparrow$ & SSIM $\uparrow$ & LPIPS $\downarrow$ && Flow EPE $\downarrow$  & PSNR $\uparrow$ & SSIM $\uparrow$ & LPIPS $\downarrow$ \\
    \midrule
    Diffusion                    & 1.79 & 30.18 & 0.9526 & 0.049 && 0.04 & 41.12 & 0.9902 & 0.008 \\
    Xu~\etal~\cite{Xu-CVPR-DFVI} & 2.01 & 27.17 & 0.9216 & 0.085 && 0.26 & 38.88 & 0.9882 & 0.009 \\
    Ours                         & 1.63 & 30.91 & 0.9564 & 0.048 && 0.03 & 41.89 & 0.9910 & 0.007 \\
    \bottomrule
    \end{tabular}
    }
}
\end{table}

We compare the proposed gradient propagation process with color propagation (used in \cite{Huang-SIGGRAPH-temporally,Xu-CVPR-DFVI}).
\figref{gradient} shows a visual comparison.
When filling the missing region with directly propagated colors, the result contains visible seams due to color differences in different source frames (\figref{gradient}a), which are removed when operating in the gradient domain (\figref{gradient}c).
\tabref{ablation}(a) analyzes the contribution of the gradient propagation quantitatively.

\heading{Non-local temporal neighbors.}
We study the effectiveness of the non-local temporal neighbors.
\tabref{ablation}(a) shows the quantitative comparisons. 
The overall quantitative improvement is somewhat subtle because, in many simple scenarios, the forward/backward flow neighbors are sufficient for propagating the correct content.
In challenging cases, the use of non-local neighbors helps substantially reduce artifacts when both forward and backward (transitively connected) flow neighbors are incorrect due to occlusion or not available. 
\figref{nonlocal} shows such an example.
Using non-local neighbors enables us to transfers the correct contents from temporally distant frames.

\heading{Edge-guided flow completion.} 
We evaluate the performance of completing the flow field with different methods. 
In \figref{flow}, we show two examples of flow completion results using diffusion (essentially Equation~\ref{eq:flow_completion} without edge guidance), a trained flow completion network~\cite{Xu-CVPR-DFVI}, and our proposed edge-guided flow completion.
The diffusion-based method maximizes smoothness in the flow field everywhere and thus cannot create motion boundaries.
The learning-based flow completion network \cite{Xu-CVPR-DFVI} fails to predict a smooth flow field and sharp flow edges.  
In contrast, the proposed edge-guided flow completion fills the missing region with a piecewise-smooth flow and no visible seams along the hole boundary.
\tabref{ablation}(b) reports the endpoint error (EPE) between the pseudo ground truth flow (i.e., flow computed from the original, uncorrupted videos using FlowNet2) and the completed flow.
The results show that the proposed flow completion achieves significantly lower EPE errors than diffusion and the trained flow completion network~\cite{Xu-CVPR-DFVI}.
As a result, our proposed flow completion method helps improve the quantitative results.

\subsection{Limitations}
\newlength\ftlm
\setlength\ftlm{4cm}
\newlength\ftln
\setlength\ftln{0.4mm}
\newlength\ftlo
\setlength\ftlo{-0.8mm}

\begin{figure}[t]
\centering%
\parbox[t]{\ftlm}{\centering%
  \fbox{\includegraphics[width=\ftlm]{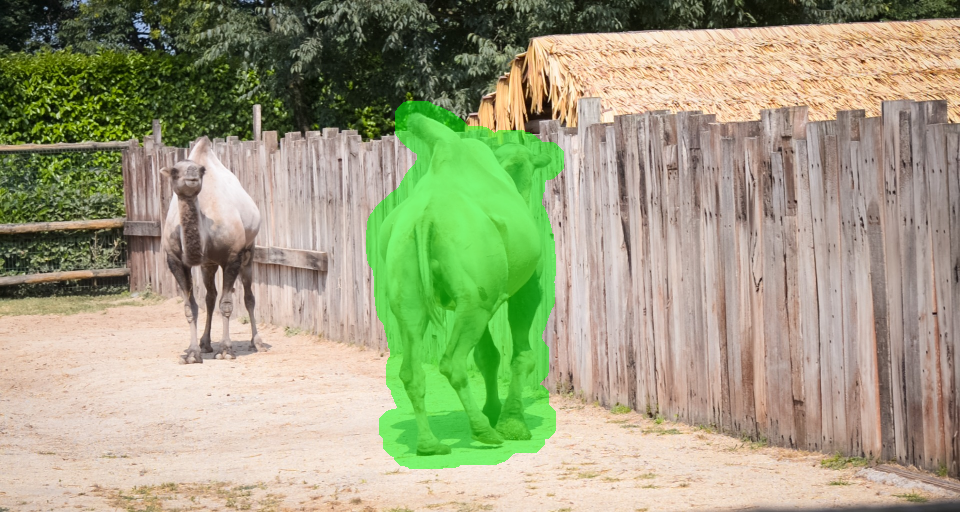}}\\%
  \fbox{\includegraphics[width=\ftlm]{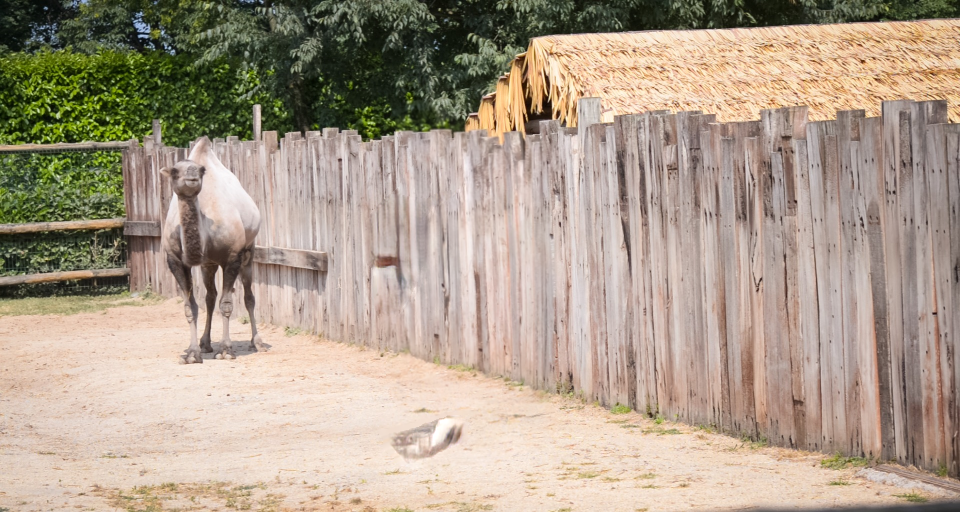}}\\%
	\small Large stationary hole}%
\hfill%
\parbox[t]{\ftlm}{\centering%
  \fbox{\includegraphics[width=\ftlm]{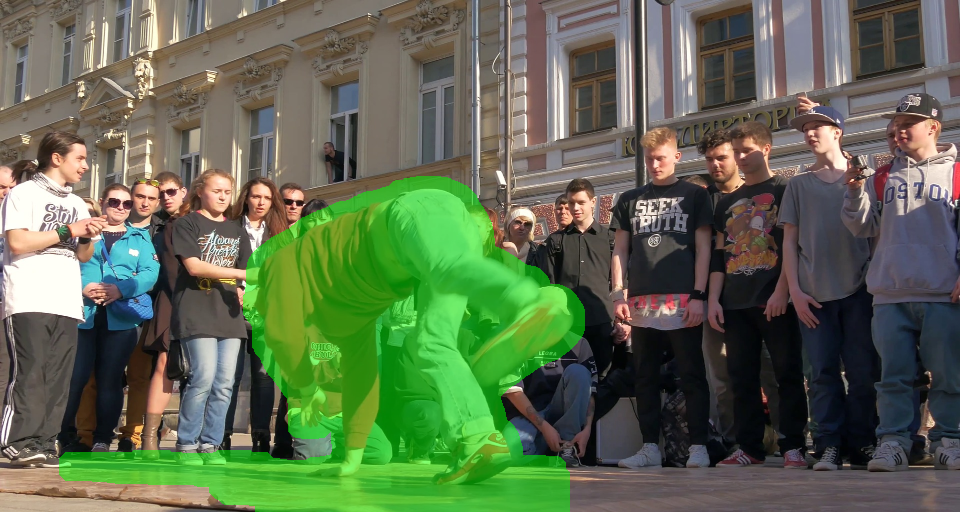}}\\%
  \fbox{\includegraphics[width=\ftlm]{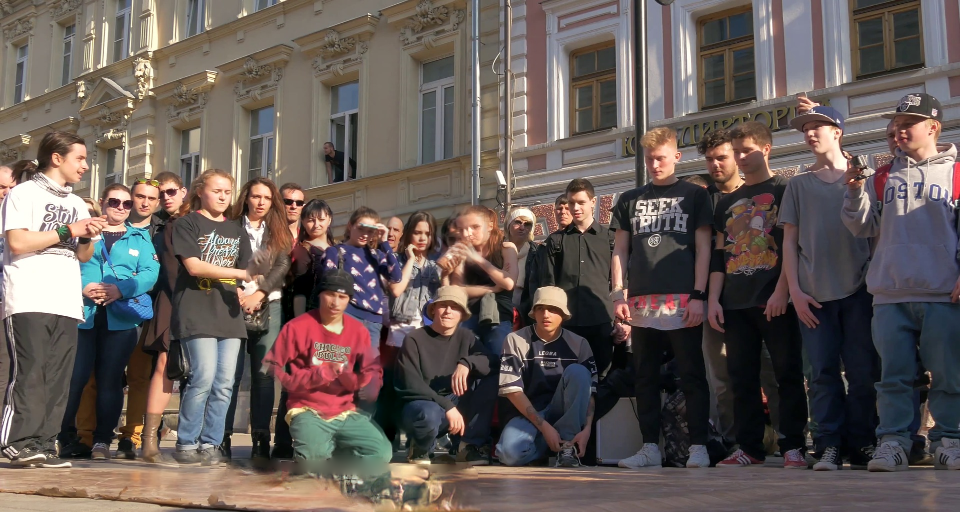}}\\%
	\small Semantic structure}%
\hfill%
\parbox[t]{\ftlm}{\centering%
  \fbox{\includegraphics[width=\ftlm]{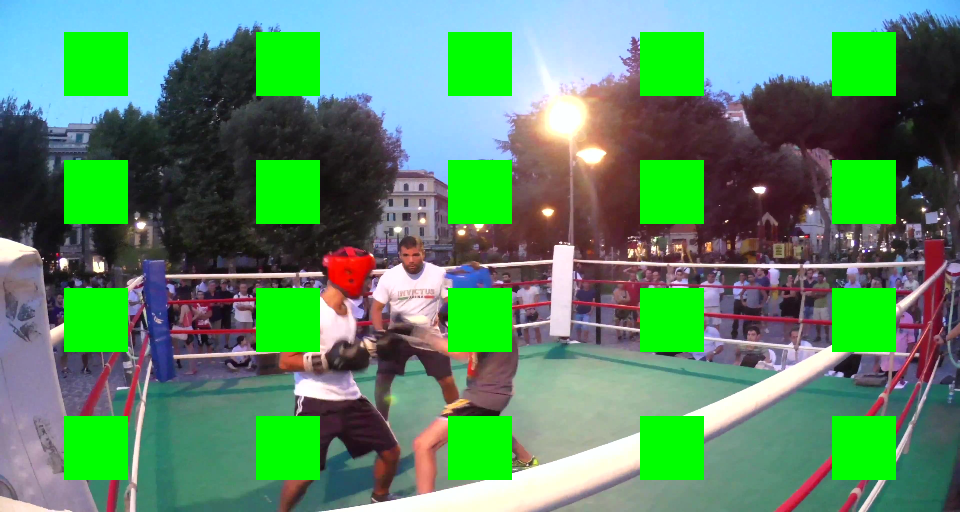}}\\%
  \fbox{\includegraphics[width=\ftlm]{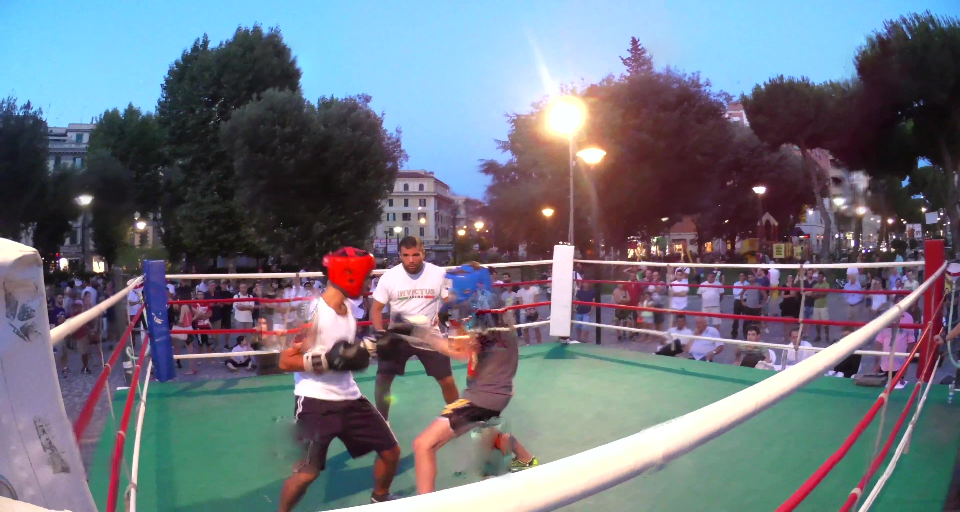}}\\%
	\small Fast motion}%

\caption{\textbf{Failure cases.}
Left, middle: hallucinated content in large missing regions (i.e., not filled by propagation) is sometimes not plausible. Right: fast motion might lead to poorly estimated flow, which results in a poor color completion.
}
\label{fig:failure}
\end{figure}

\heading{Failure results.} Video completion remains a challenging problem. 
We show and explain several failure cases in \figref{failure}.

\heading{Processing speed.} Our method runs at $0.12$ fps, which is comparable to other flow-based methods.
End-to-end models are relatively faster, \eg Lee \etal \cite{Lee-ICCV-CPNet} runs at $0.405$ fps, but with worse performance. 
We acknowledge our slightly slower running time to be a weakness.

\subsection{Negative results}
\label{sec:negative}

We explored several alternatives to our design choices to improve the quality of our video completion results.
Unfortunately, these changes either ended up degrading performance or not producing clear improvement.

\heading{Flow completion network.}
As many CNN-based methods have shown impressive results on the task of image completion, using a CNN for flow completion seems a natural approach.
We modified and experimented with several inpainting architectures, including partial conv~\cite{Liu-ECCV-Partialconv} and EdgeConnect~\cite{Nazeri-ICCVW-Edgeconnect} for learning to complete the missing flow (by training on flow fields extracted from a large video dataset~\cite{Kay-kinetics}). 
However, we found that in both cases, the network fails to generalize to unseen video sequences and produce visible seams along the hole boundaries. 
%

\heading{Learning-based fusion.}
We explored using a
U-Net based model for learning the weights for fusing the candidate (\secref{candidate_fusion}).
Our model takes a forward-backward consistency error maps and the validity mask as inputs and predict the fusion weights so that the fused gradients are as similar to the ground truth gradients as possible. 
However, we did not observe a clear improvement from this learning-based method over the hand-crafted weights.



\clearpage
%
%
\bibliographystyle{splncs04}
\bibliography{references}
\end{document}


\pagestyle{headings}
\mainmatter
\def\ECCVSubNumber{1715}  

\title{Flow-edge Guided Video Completion \\
Supplementary Material}

\author{
Chen Gao\inst{1} \and
Ayush Saraf\inst{2} \and
Jia-Bin Huang\inst{1} \and
Johannes Kopf\inst{2}
}
\authorrunning{C. Gao et al.}
\institute{${}^{1}$ Virginia Tech \quad ${}^{2}$ Facebook
}

\maketitle

\section*{Overview}
In this supplementary document, we provide additional implementation details and results to complement the main manuscript.

\begin{enumerate}
\item We summarize the complete pipeline of our algorithm in pseudo-code in \algref{alg:algorithm}.
\item We show the runtime analysis and profiling to analyze the speed of our algorithm.
\item We describe the training details for the edge completion network. 
\item We present additional visual examples of the ablation study, highlighting the effectiveness of our design choices. 
\item We provide detailed per-sequence results in terms of PSNR, SSIM, and LPIPS on the DAVIS dataset.
\end{enumerate}

\colorlet{pink}{red!60}
\colorlet{cyan}{cyan!70}
\colorlet{azure}{green!50}

\newpage
\section{Algorithm}
We show our method pipeline in \href{run:./algorithm_illustration.mp4}{algorithm\_illustration.mp4}.
We summarize our complete pipeline in \algref{alg:algorithm}. 
Our pipeline consists of three main components: \textcolor{red!90}{flow prediction}, \textcolor{cyan!90}{edge-guided flow completion}, and \textcolor{green!90}{video completion}. 
%
We will release the pre-trained flow-edge completion model as well as the source code to facilitate future research.

\begin{algorithm}[ht]
\SetAlgoLined
\SetAlgoNoEnd
\textbf{Input:} Color frames $\bm{I}_1 $\dots$ \bm{I}_{n}$, mask frames $\bm{M}_1 $\dots$ \bm{M}_n$.\\
\textbf{Output:} Completed frames $\bm{I}_1 $\dots$ \bm{I}_n$ (updated in place).\\
\tikzmk{A}\For{$\emph{every frame~} i \in 1 $\dots$ n$}{
Compute local and non-local optical flow $\bm{F}_{\ij},$ \\
$j \in \{ i-1, i+1, 1, \left\lceil n/2 \right\rceil, n \}$
(Equations \red{1} and \red{2}).
}
\tikzmk{B}
\boxit{pink}
\tikzmk{A}\For{\emph{each computed flow field~} $\bm{F}_{\ij}$}{
Extract flow edges $\bm{E}_{\ij}$ using Canny edge detector \cite{Canny-1986:Edge}.\\
Complete flow edges $\tilde{\bm{E}}_{\ij}$ using EdgeConnect \cite{Nazeri-ICCVW-Edgeconnect} edge model.\\
Complete flow $\tilde{\bm{F}}_{\ij}$ with edge guidance (Equation \red{3}).\\
Compute flow error $\tilde{\bm{D}}_{\ij}$ (Equation \red{4}).
}
\tikzmk{B}
\boxit{cyan}
\tikzmk{A}\While{$\emph{any missing pixels exist in~} \bm{M}_1 $\dots$ \bm{M}_n$}{
\For{$\emph{every frame~} i \in 1 $\dots$ n$}{
\label{marker}
Obtain temporal neighbors through propagation.\\
Fuse gradient images $\tilde{\bm{G}}_{x,i}$ and $\tilde{\bm{G}}_{y,i}$ (Equation \red{7}).\\
Reconstruct color image $\tilde{\bm{I}}_i$ (Equation \red{8}).\\
Update mask $\bm{M}_i\p = 0$, where $\left| N\p \right| \geq 1$.
}
Select frame $\tilde{\bm{I}}_f$ with most remaining missing pixels.\\
Complete $\tilde{\bm{I}}_f$ with DeepFill~\cite{Yu2018-Generative}.\\
Set $\bm{M}_f = 1$ (all pixels in this frame).\\
Set $\bm{I} = \tilde{\bm{I}}$.}
\tikzmk{B}
\boxit{azure}
\caption{Summary of our video completion algorithm.}
\label{alg:algorithm}
\end{algorithm}


\begin{table}[ht]
    \scriptsize
    \caption{
    \textbf{Running time analysis.} We report the running time for each component of our method on the ``CAMEL'' video sequence under the object removal setting. The resolution is $960 \times 512$.
    }
    \label{tab:running_time}
    \centering
    \resizebox{0.7\columnwidth}{!}{
    \begin{tabular}{l|l|c}
    \toprule
    & Component & Time (min.) \\
    \midrule
    \multirow{3}{*}{Flow completion} & \cellcolor{red!20} Flow prediction                 & \cellcolor{red!20}1.20 \\ 
                                     & \cellcolor{cyan!20} Edge extraction and completion & \cellcolor{cyan!20}0.45 \\ 
                                     & \cellcolor{cyan!20} Edge-guided flow completion    & \cellcolor{cyan!20}4.20  \\  
    \cline{0-0}
    \multirow{3}{*}{Video completion}& \cellcolor{green!15} Temporal propagation         & \cellcolor{green!15}4.31 \\ %
                                     & \cellcolor{green!15} Spatial inpainting           & \cellcolor{green!15}0.10 \\ %
                                     & \cellcolor{green!15} Poisson blending             & \cellcolor{green!15}2.29 \\ %
    \midrule
    Total &                          & 12.55 \\ %
    \bottomrule
    \end{tabular}
    }
\end{table}

\begin{figure}[th]
\newlength\ftfc
\setlength\ftfc{2.4cm}
\newlength\ftfd
\setlength\ftfd{0.8mm}
\newlength\ftfe
\setlength\ftfe{-0.0mm}
\parbox[t]{\ftfc}{\centering%
\fbox{\includegraphics[width=\ftfc]{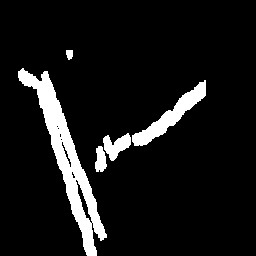}}\\%
\fbox{\includegraphics[width=\ftfc]{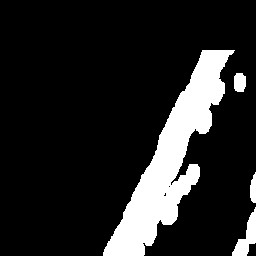}}\\%
\fbox{\includegraphics[width=\ftfc]{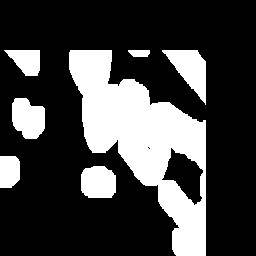}}\\%
\scriptsize (a) Mask}%
\hfill%
\parbox[t]{\ftfc}{\centering%
\fbox{\includegraphics[width=\ftfc]{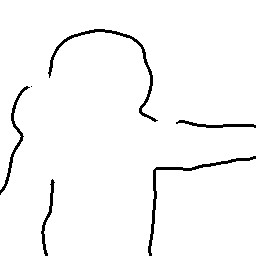}}\\%
\fbox{\includegraphics[width=\ftfc]{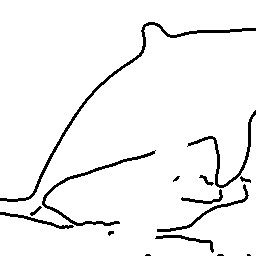}}\\%
\fbox{\includegraphics[width=\ftfc]{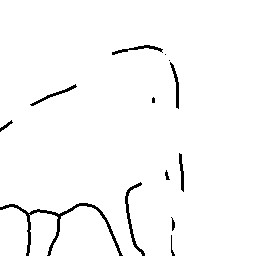}}\\%
\scriptsize (b) Input edge}%
\hfill%
\parbox[t]{\ftfc}{\centering%
\fbox{\includegraphics[width=\ftfc]{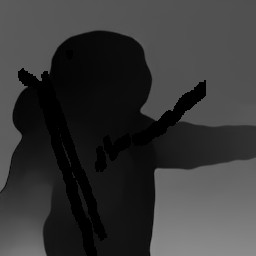}}\\%
\fbox{\includegraphics[width=\ftfc]{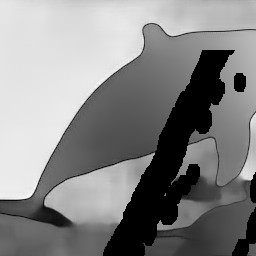}}\\%
\fbox{\includegraphics[width=\ftfc]{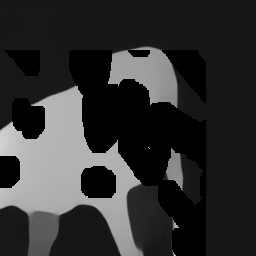}}\\%
\scriptsize (c) Input flow mag.}%
\hfill%
\parbox[t]{\ftfc}{\centering%
\fbox{\includegraphics[width=\ftfc]{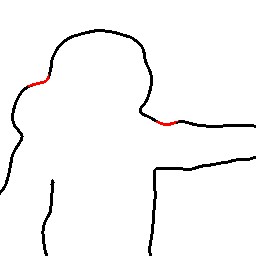}}\\%
\fbox{\includegraphics[width=\ftfc]{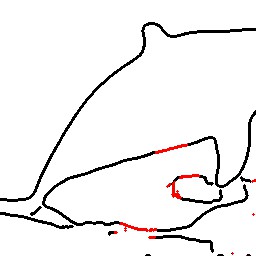}}\\%
\fbox{\includegraphics[width=\ftfc]{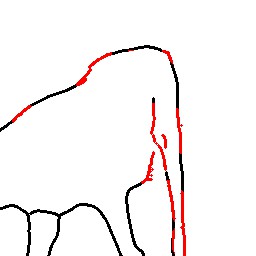}}\\%
\scriptsize (d) Output edge}%
\hfill%
\parbox[t]{\ftfc}{\centering%
\fbox{\includegraphics[width=\ftfc]{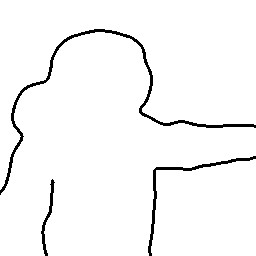}}\\%
\fbox{\includegraphics[width=\ftfc]{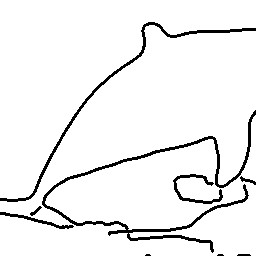}}\\%
\fbox{\includegraphics[width=\ftfc]{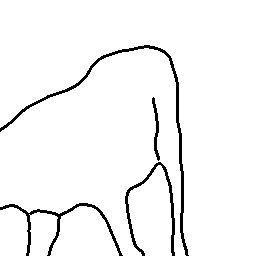}}\\%
\scriptsize (e) Ground truth}%
\\
$\underbracket[1pt][2.0mm]{\hspace{0.6\textwidth}}_%
    {\substack{\vspace{-3.0mm}\\\colorbox{white}{~~Input~~}}}$\vspace{1mm}\\%
\caption{\textbf{Flow edge completion.} Our flow edge completion network takes the mask, the corrupted flow edge map and the corrupted flow magnitude image as input, and complete the flow edge.
}
\label{fig:edge}
\end{figure}

\section{Runtime analysis and profiling}
Following~\cite{Huang-SIGGRAPH-temporally}, we also show the detailed running time analysis of our method in \tabref{running_time}. 
We report the time for each component of our method on the ``CAMEL'' video sequence under the object removal setting. 
The resolution is $960 \times 512$. There are $10721523$ pixels being removed, which is $9.1\%$ of the total pixels.
Our method runs at $7.2$ frames per minute.

\section{Training details}
The only trainable component in our method is the flow edge completion network. 
We build our flow edge completion network upon the publicly available official implementation of EdgeConnect~\cite{Nazeri-ICCVW-Edgeconnect} edge model\footnote{\url{https://github.com/knazeri/edge-connect}}.
We load weights pretrained on the Places2 dataset \cite{zhou2017places}, and then finetune on 60 sequences in DAVIS \texttt{2017-test-dev} and \texttt{2017-test-challenge} for three epochs.

Starting from the predicted flow between adjacent frames $i$ and $j$, $\bm{F}_{i \rightarrow j}$, we first calculate the flow magnitude image. 
We use the Canny edge detector \cite{Canny-1986:Edge} to extract a flow edge map $\bm{E}_{\ij}$ from the flow magnitude image. 
We use the following parameters for the Canny edge detector~\cite{Canny-1986:Edge}: Gaussian $\sigma = 1$, low threshold $0.1$, high threshold $0.2$.
We randomly choose a mask from NVIDIA Irregular Mask Dataset testing split and resize it to $256 \times 256$.\footnote{\url{https://www.dropbox.com/s/01dfayns9s0kevy/test_mask.zip?dl=0}}
We crop the flow edge map $\bm{E}_{\ij}$ and the corresponding flow magnitude images to $256 \times 256$, and corrupt them with the mask.
The input to the flow edge completion network is the mask (\figref{edge}a), the corrupted flow edge map (\figref{edge}b) and the corrupted flow magnitude image (\figref{edge}c).
We train the network to complete the flow edge using batches of 8 randomly cropped $256 \times 256$ patches. 

Note that our edge completion network does \emph{not} receive any additional information regarding the stationary mask with a uniform grid of $5 \times 4$ square blocks during training.





\newlength\ftgd
\setlength\ftgd{4.0cm}
\newlength\ftge
\setlength\ftge{0.0mm}
\begin{figure}[t]

\centering%
\parbox[t]{\ftgd}{\centering%
\includegraphics[width=\ftgd]{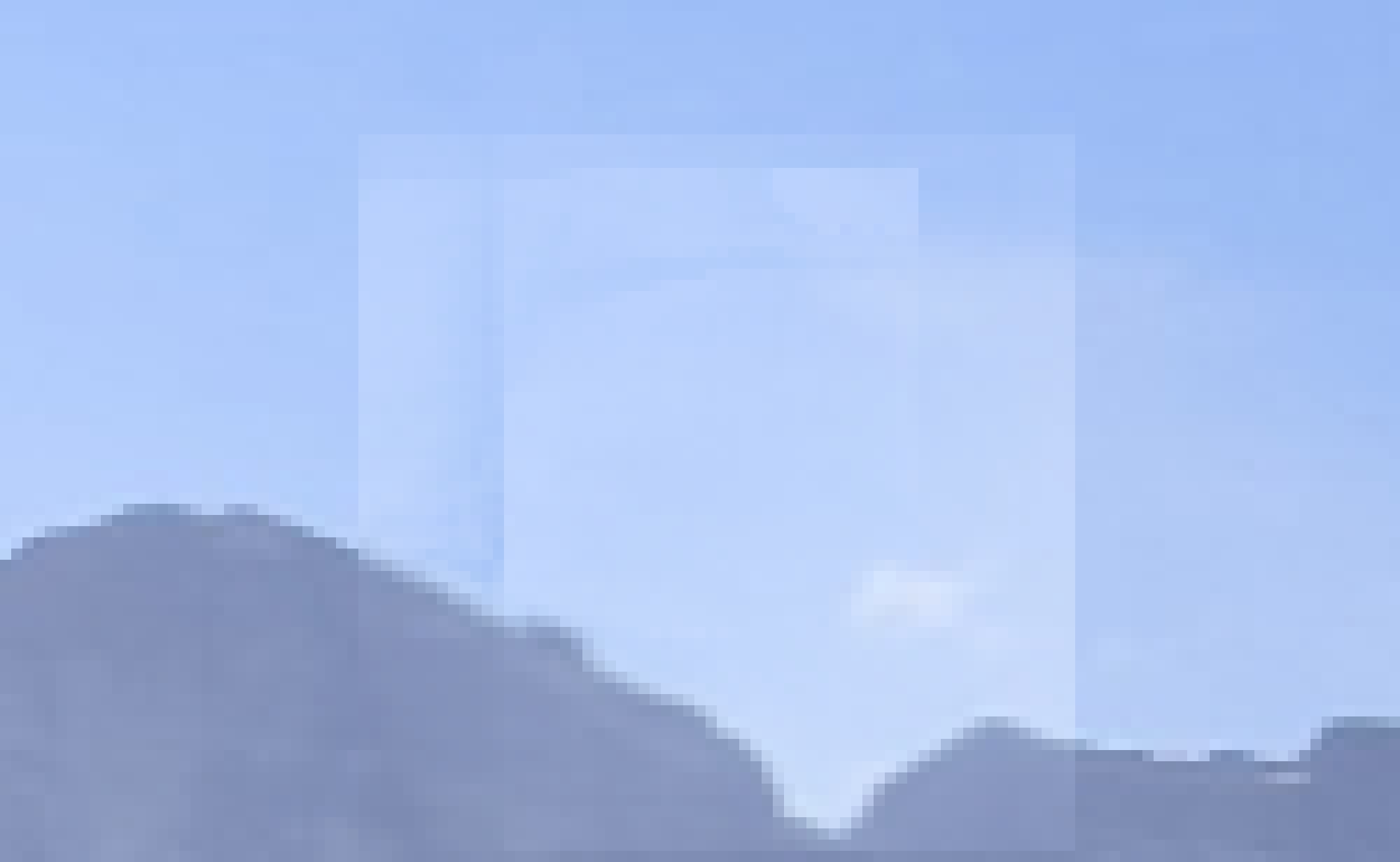}\\%
\includegraphics[width=\ftgd]{fig/ablation_gradient/bear_RGB_00000_crop_mag.jpg}\\%
\includegraphics[width=\ftgd]{fig/ablation_gradient/swing_RGB_00008_crop_mag.jpg}\\%
\includegraphics[width=\ftgd]{fig/ablation_gradient/bmx-bumps_RGB_00001_crop_mag.jpg}
\small (a) Color propagation}%
\hfill%
\parbox[t]{\ftgd}{\centering%
\includegraphics[width=\ftgd]{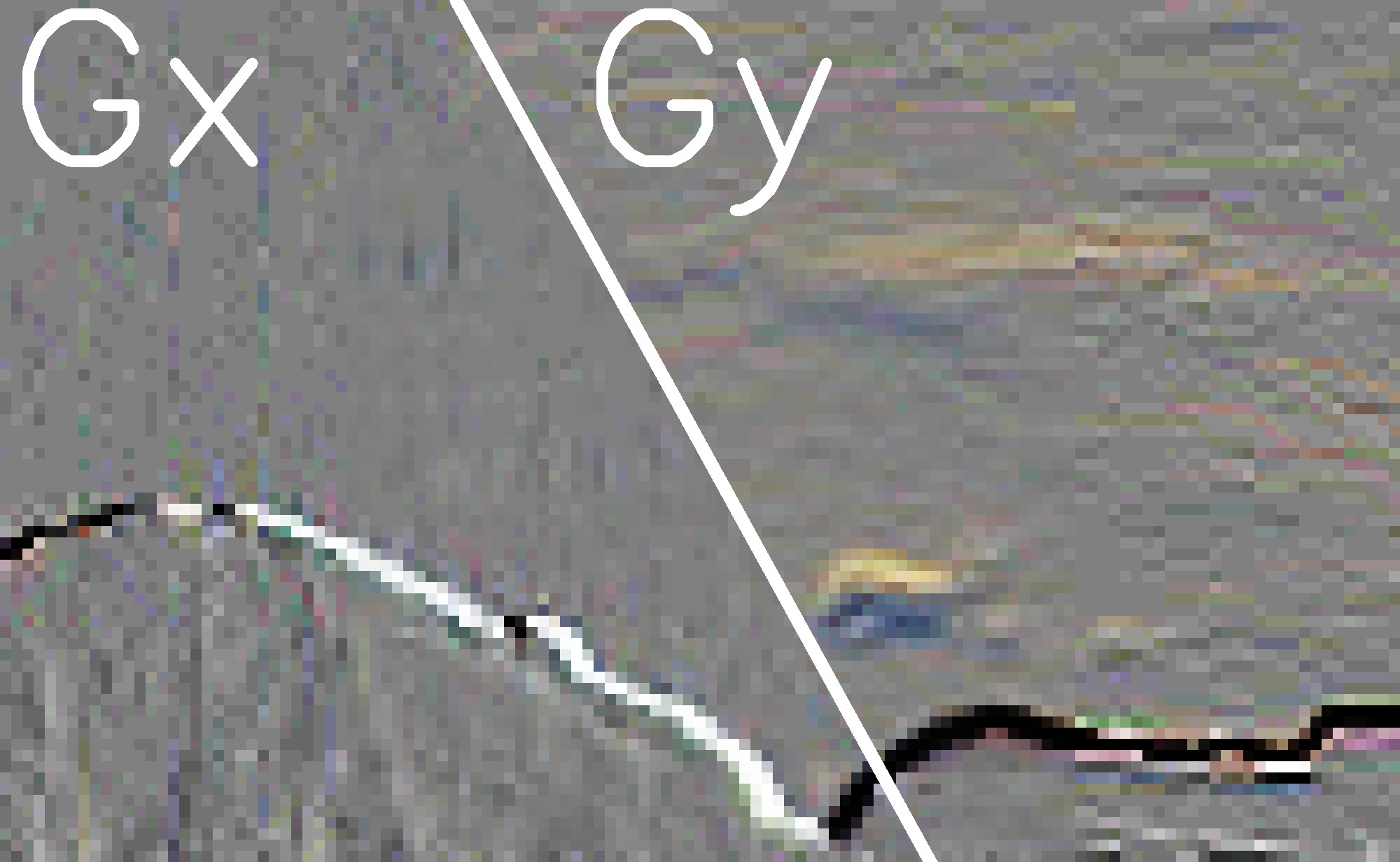}\\%
\includegraphics[width=\ftgd]{fig/ablation_gradient/bear_xy_00000_crop_mag.jpg}\\%
\includegraphics[width=\ftgd]{fig/ablation_gradient/swing_xy_00008_crop_mag.jpg}\\%
\includegraphics[width=\ftgd]{fig/ablation_gradient/bmx-bumps_xy_00001_crop_mag.jpg}
\small (b) Propagated x/y gradient}%
\hfill%
\parbox[t]{\ftgd}{\centering%
\includegraphics[width=\ftgd]{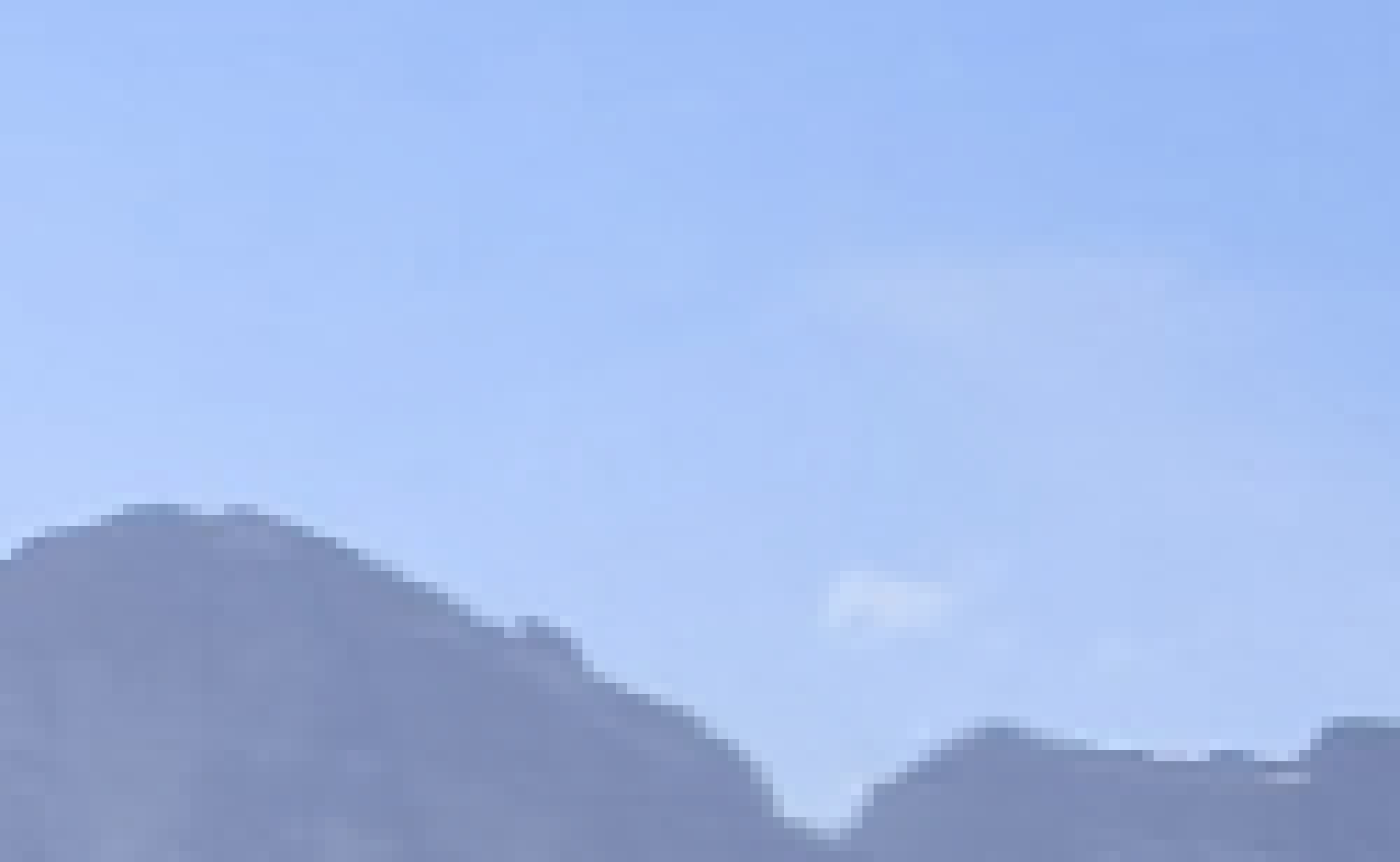}\\%
\includegraphics[width=\ftgd]{fig/ablation_gradient/bear_Gradient_00000_crop_mag.jpg}\\%
\includegraphics[width=\ftgd]{fig/ablation_gradient/swing_Gradient_00008_crop_mag.jpg}\\%
\includegraphics[width=\ftgd]{fig/ablation_gradient/bmx-bumps_Gradient_00001_crop_mag.jpg}
\small (c) Reconstruction}%

\caption{\textbf{Gradient domain reconstruction.}
Previous methods operate directly in the color domain, which results in visible seams in the completed video (a). We propagate in the gradient domain (b), and reconstruct the results via Poisson reconstruction (c).
}
\label{fig:gradient}
\end{figure}
\newlength\ftgf
\setlength\ftgf{6.05cm}
\newlength\ftgg
\setlength\ftgg{0.0mm}
\begin{figure}[t]
\centering%
\parbox[t]{\ftgf}{\centering%
\includegraphics[width=\ftgf]{fig/ablation_nonlocal/bus_00032_wo_nl.jpg}\\%
\includegraphics[width=\ftgf]{fig/ablation_nonlocal/drift-straight_00021_wo_nl.jpg}\\%
\includegraphics[width=\ftgf]{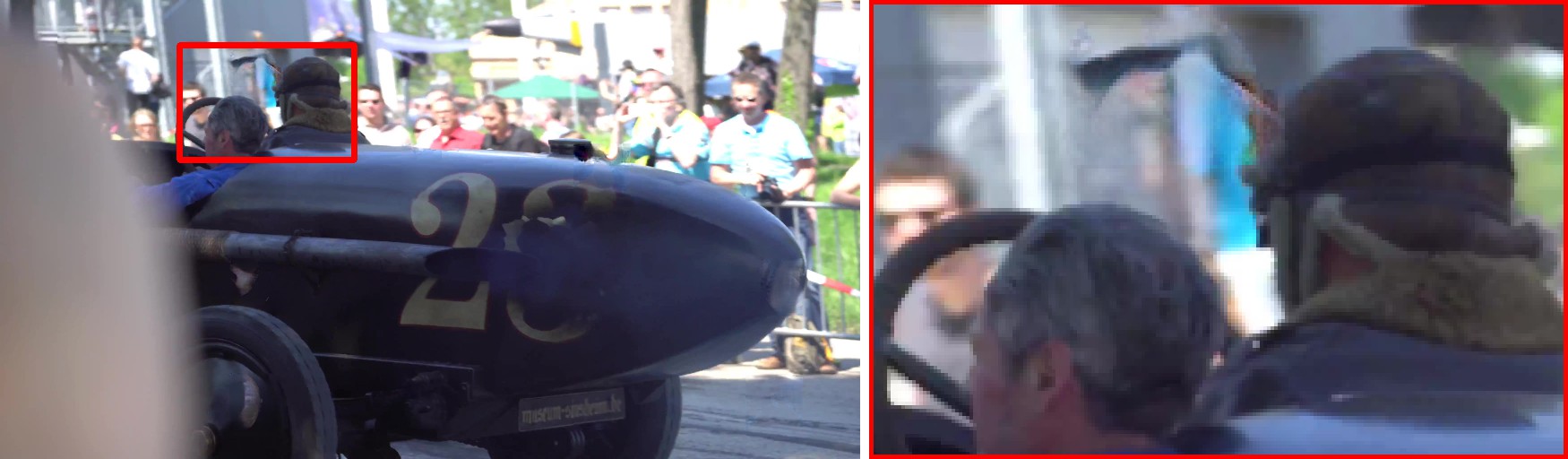}\\%
Without non-local neighbors}%
\hfill%
\parbox[t]{\ftgf}{\centering%
\includegraphics[width=\ftgf]{fig/ablation_nonlocal/bus_00032_w_nl.jpg}\\%
\includegraphics[width=\ftgf]{fig/ablation_nonlocal/drift-straight_00021_w_nl.jpg}\\%
\includegraphics[width=\ftgf]{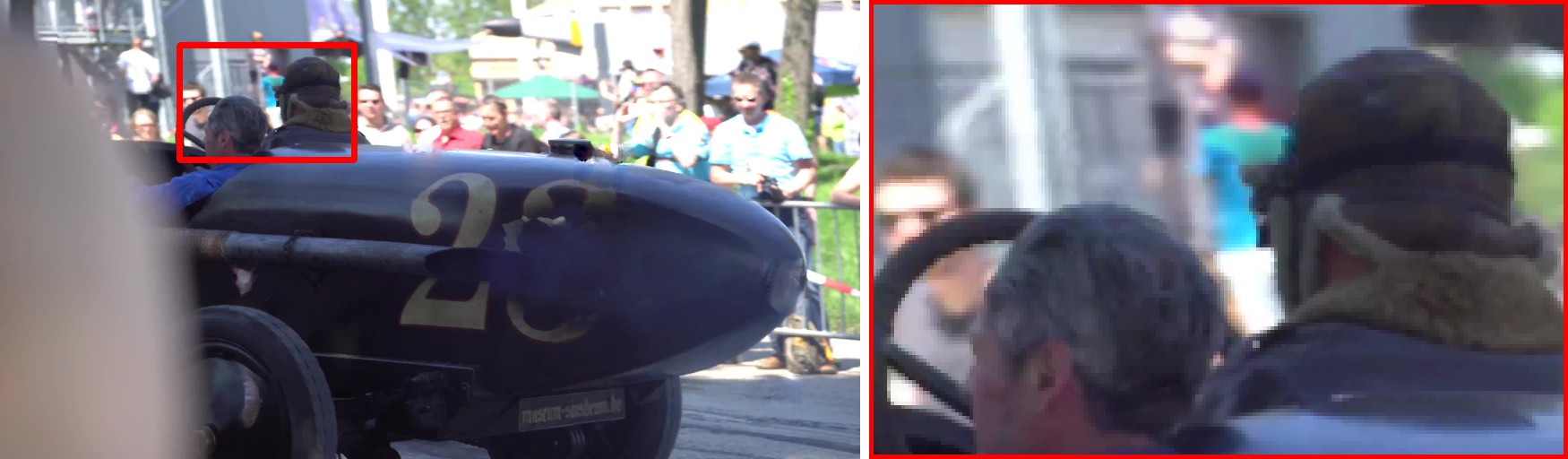}\\%
With non-local neighbors}%
\caption{\textbf{Non-local temporal neighbor ablation.}
Video completion results \emph{with} and \emph{without} non-local temporal neighbors.
The result without non-local neighbors (left) does not recover well from the lack of well-propagated content.
}
\label{fig:nonlocal}
\end{figure}
\begin{figure}[t]
\centering%
\setlength\ftfc{3.0cm}
\setlength\ftfd{0.0mm}
\setlength\ftfe{-0.8mm}
\parbox[t]{\ftfc}{\centering%
  \fbox{\includegraphics[width=\ftfc]{fig/ablation_flow/input.jpg}}\\%
 \fbox{\includegraphics[width=\ftfc]{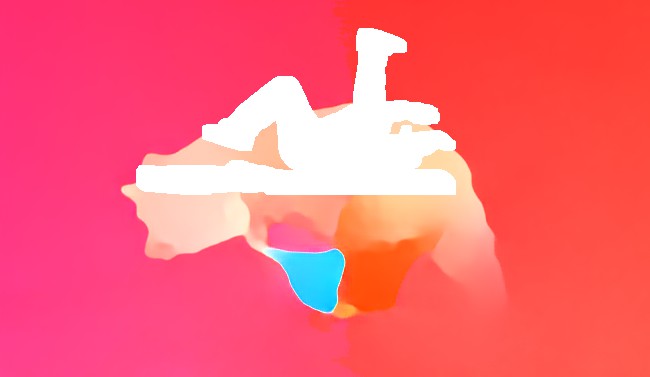}}\\%
	\footnotesize Input}%
\hfill%
\parbox[t]{\ftfc}{\centering%
  \fbox{\includegraphics[width=\ftfc]{fig/ablation_flow/ours_wo_edge.jpg}}\\%
 \fbox{\includegraphics[width=\ftfc]{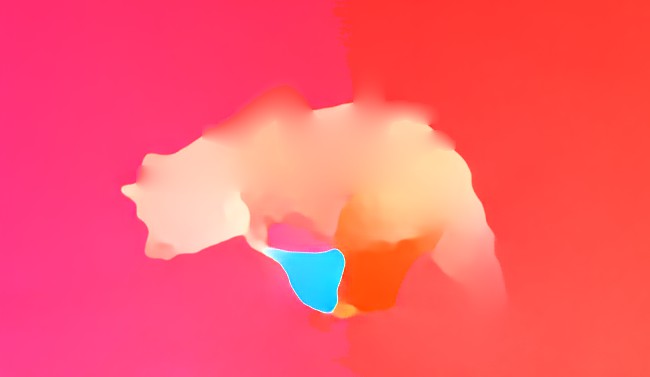}}\\%
	\footnotesize Diffusion}%
\hfill%
\parbox[t]{\ftfc}{\centering%
  \fbox{\includegraphics[width=\ftfc]{fig/ablation_flow/DFVI.jpg}}\\%
 \fbox{\includegraphics[width=\ftfc]{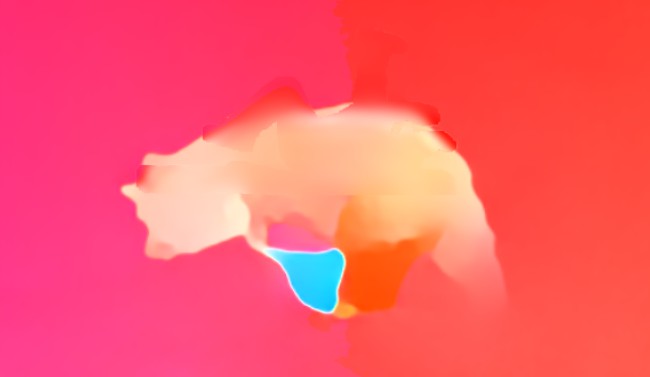}}\\%
	\footnotesize Xu~\etal~\cite{Xu-CVPR-DFVI}}%
\hfill%
\parbox[t]{\ftfc}{\centering%
  \fbox{\includegraphics[width=\ftfc]{fig/ablation_flow/ours_w_edge.jpg}}\\%
 \fbox{\includegraphics[width=\ftfc]{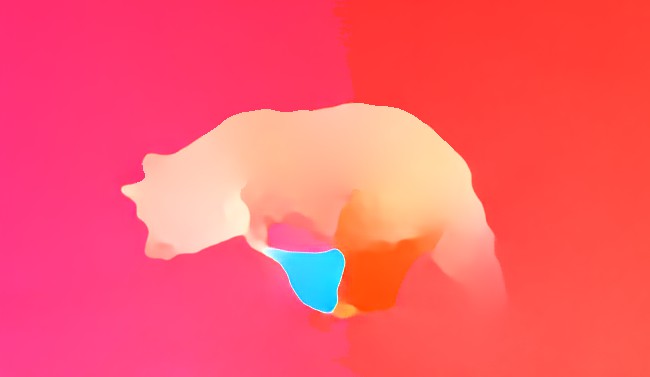}}\\%
	\footnotesize Ours}%
\caption{\textbf{Flow completion.} Comparing different methods for flow completion. Our method has better ability to retain the piecewise-smooth nature of flow fields (sharp motion boundaries, smooth everywhere else) than the other two methods.
}
\label{fig:flow}
\end{figure}

\section{Additional visual examples of the ablation study}

In this section, we show additional visual examples of the ablation study to highlight the effectiveness of our design choices.

\heading{Gradient domain processing.} 
We compare the proposed gradient propagation process with color propagation (used in \cite{Huang-SIGGRAPH-temporally,Xu-CVPR-DFVI}).
\figref{gradient} shows a visual comparison.
When filling the missing region with directly propagated colors, the result contains visible seams due to color differences in different source frames (\figref{gradient}a).
Our method operates in the gradient domain  and does not suffers from such artifacts (\figref{gradient}c).

\heading{Non-local temporal neighbors.}
We study the effectiveness of the non-local temporal neighbors.
\figref{nonlocal} shows such an example.
Using non-local neighbors enables us to transfers the correct contents from temporally distant frames.

\heading{Edge-guided flow completion.} 
We evaluate the performance of completing the flow field with different methods. 
In \figref{flow}, we show examples of flow completion results using diffusion, a trained flow completion network~\cite{Xu-CVPR-DFVI}, and our proposed edge-guided flow completion.
The diffusion-based method maximizes smoothness in the flow field everywhere and thus cannot create sharp motion boundaries.
The learning-based flow completion network \cite{Xu-CVPR-DFVI} fails to predict a smooth flow field and sharp flow edges.
In contrast, the proposed edge-guided flow completion fills the missing region with a piecewise-smooth flow and no visible seams along the hole boundary.

\begin{figure}[ht]
\centering
\includegraphics[width=\linewidth]{fig/per_sequence_v2.pdf}
\caption{\textbf{Per-sequence PSNR, SSIM and LPIPS on DAVIS under the stationary masks inpainting setting.}
}
\label{fig:per_sequence}
\end{figure}
\section{Per-sequence results on the DAVIS dataset}
We show the detailed per-sequence results in terms of PSNR, SSIM, and LPIPS under stationary screen-space masks setting in \figref{per_sequence}.
Our proposed method improves the performance over state-of-the-art methods for most of the video sequences, under all three metrics.

\clearpage
\bibliographystyle{splncs04}
\bibliography{references}